\documentclass[11pt]{article}

\usepackage[final]{acl}

\usepackage{times}
\usepackage{latexsym}

\usepackage[T1]{fontenc}

\usepackage[utf8]{inputenc}

\usepackage{microtype}

\usepackage{inconsolata}

\usepackage{graphicx}

\usepackage{hyperref}
\newcommand{\methodname}{\texttt{HumanLLM}\xspace}

\usepackage{colortbl}
\usepackage{booktabs}
\usepackage{enumitem}
\usepackage{xspace}
\usepackage{array}
\usepackage{multirow}
\usepackage{subcaption}

\title{{\methodname}: Benchmarking and Improving LLM Anthropomorphism via Human Cognitive Patterns}

\author{
  \textbf{Xintao Wang}\textsuperscript{1}\thanks{\ \,Equal contribution.}, 
  \textbf{Jian Yang}\textsuperscript{1}\footnotemark[1], 
  \textbf{Weiyuan Li}\textsuperscript{1}, 
  \textbf{Rui Xie}\textsuperscript{1}, 
  \textbf{Jen-tse Huang}\textsuperscript{3}, 
  \textbf{Jun Gao}\textsuperscript{2}, \\
  \textbf{Shuai Huang}\textsuperscript{2}, 
  \textbf{Yueping Kang}\textsuperscript{2}, 
  \textbf{Yuanli Guo}\textsuperscript{1}, 
  \textbf{Hongwei Feng}\textsuperscript{1}\thanks{\ \,Corresponding authors.}, 
  \textbf{Yanghua Xiao}\textsuperscript{1}\footnotemark[2]
  \\[0.3em]
  \textsuperscript{1}Fudan University \quad
  \textsuperscript{2}Hello Group \quad
  \textsuperscript{3}Johns Hopkins University \\[0.3em]
  \texttt{\{xtwang21, 24210240375, 25210980069, 25210980167\}@m.fudan.edu.cn} \\
  \texttt{\{hwfeng, shawyh, guoyuanli\}@fudan.edu.cn} \\
  \texttt{gaojun55@gmail.com}, 
  \texttt{\{huang.shuai, kang.yueping\}@hellogroup.com}, 
  \texttt{jhuan236@jh.edu}
}

\begin{document}
\maketitle
\begin{abstract}
Large Language Models (LLMs) have demonstrated remarkable capabilities in reasoning and generation, serving as the foundation for advanced persona simulation and Role-Playing Language Agents (RPLAs). However, achieving authentic alignment with human cognitive and behavioral patterns remains a critical challenge for these agents. 
We present \methodname, a framework treating psychological patterns as interacting causal forces.
We construct 244 patterns from $\sim$12,000 academic papers and synthesize 11,359 scenarios where 2--5 patterns reinforce, conflict, or modulate each other, with multi-turn conversations expressing inner thoughts, actions, and dialogue.
Our dual-level checklists evaluate both individual pattern fidelity and emergent multi-pattern dynamics, achieving strong human alignment ($r=0.90$) while revealing that holistic metrics conflate simulation accuracy with social desirability.
\methodname-8B outperforms Qwen3-32B on multi-pattern dynamics despite 4$\times$ fewer parameters, demonstrating that authentic anthropomorphism requires cognitive modeling---simulating not just what humans do, but the psychological processes generating those behaviors.
Our dataset, code, and model are available at:~\url{https://github.com/YJGoodbye2024/HumanLLM}
\end{abstract}

\section{Introduction}

\begin{figure}[t]
  \centering
  \includegraphics[width=\columnwidth]{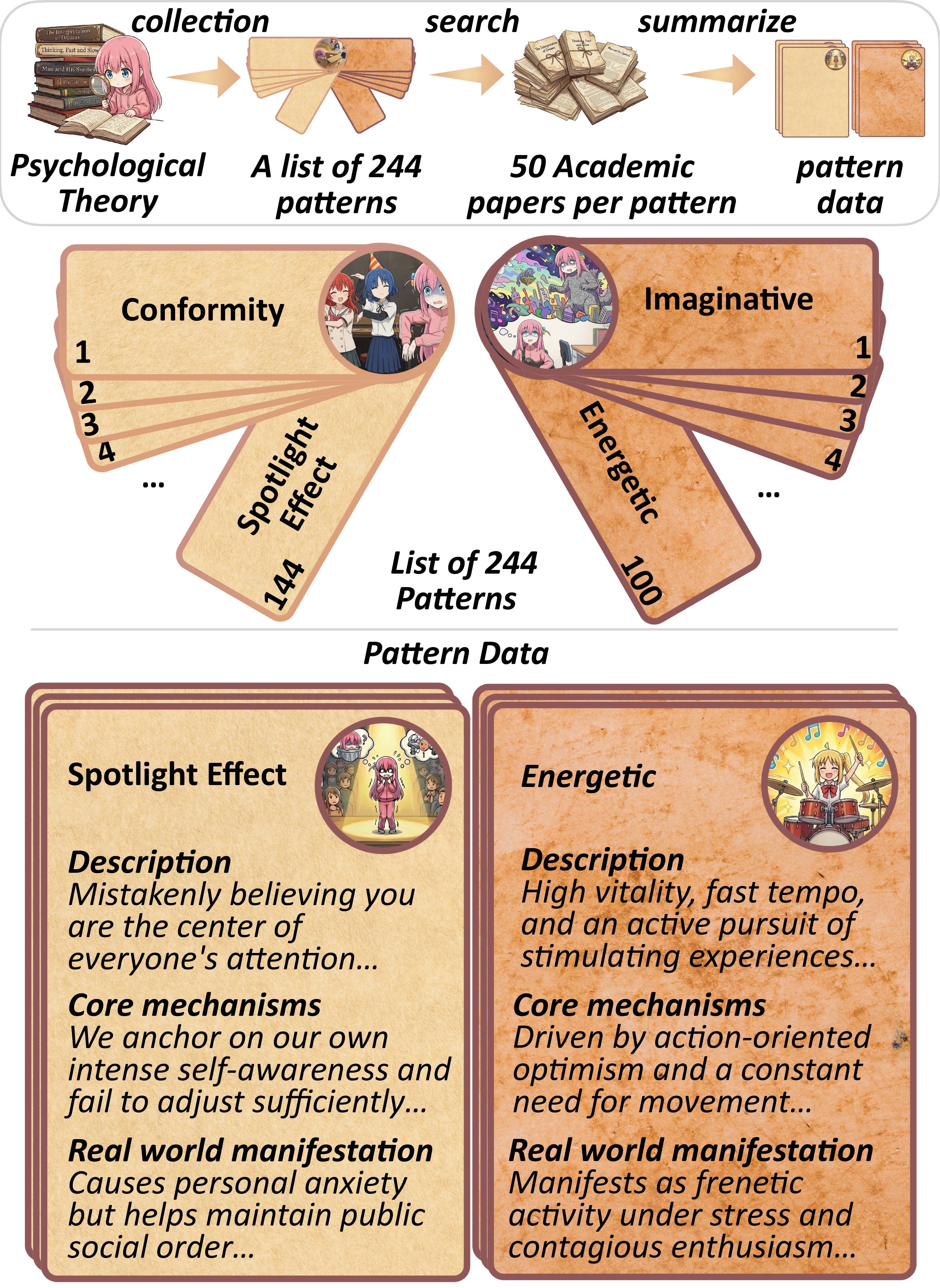}
  \caption{\textbf{Pattern Data Structure:} 144 Social-Cognitive Patterns (left) and 100 Personality Traits (right). Each pattern comprises Definition, Core Mechanisms, and Real-World Manifestations.}
  \label{fig:pattern}
\end{figure}

With the rapid scaling of training data, Large Language Models (LLMs) have achieved remarkable progress in anthropomorphism---simulating human-like characteristics and social phenomena~\citep{shanahanRolePlayLarge2023}.
Role-Playing Language Agents (RPLAs) have evolved from conceptual frameworks into practical applications~\citep{chenPersonaPersonalizationSurvey2024}, enabling digital clones~\citep{xuMINDECHORolePlayingLanguage2024}, AI companions~\citep{zhang2025riseaicompanionshumanchatbot}, and society simulation~\citep{parkGenerativeAgentsInteractive2023, zhou2025pimmur, zhou2025sotopia}.
As these applications advance, LLM anthropomorphism increasingly requires moving beyond shallow behavioral mimicry toward deeper cognitive and emotional fidelity---what we term \textbf{psychological alignment}~\citep{wangInCharacterEvaluatingPersonality2024}.

However, existing approaches model personality as isolated label-to-behavior mappings---``extroverted'' maps to ``talkative,'' ``agreeable'' maps to ``cooperative''---without capturing how multiple cognitive patterns interact to produce behavior~\citep{huang2024reliability}. We define a pattern as a psychologically documented regularity in human cognition or behavior—either a stable personality trait (e.g., "assertive") or a context-triggered social-cognitive process (e.g., "spotlight effect"). 
In reality, a talkative person may fall silent when the spotlight effect is activated; an assertive individual may yield under conformity pressure.
Human behavior emerges from the dynamic interplay of multiple patterns, not from any single trait in isolation.
Current methods---whether prompting-based~\citep{serapiogarcía2025personalitytraitslargelanguage, ng2024well}, fine-tuning-based~\citep{shaoCharacterLLMTrainableAgent2023,zhou2023characterglmcustomizingchineseconversational}, or activation steering~\citep{chen2025personavectorsmonitoringcontrolling}---all treat traits independently, leading to personality drift and the ``personality illusion'' where models report traits while behaving inconsistently~\citep{wangInCharacterEvaluatingPersonality2024,han2025personalityillusionrevealingdissociation}.

To address this, we propose \methodname framework, treating cognitive patterns not as isolated labels but as interacting causal forces. \methodname refers to the overall framework; we use '\methodname dataset' and '\methodname-8B/32B' when referring specifically to the data artifact and fine-tuned models, respectively.
Our key insight: by exposing models to scenarios where multiple patterns reinforce, compete, or conflict, models can implicitly learn multi-pattern dynamics without architectural modifications.

\begin{figure*}[t!]
  \centering
  \includegraphics[width=1.0\textwidth]{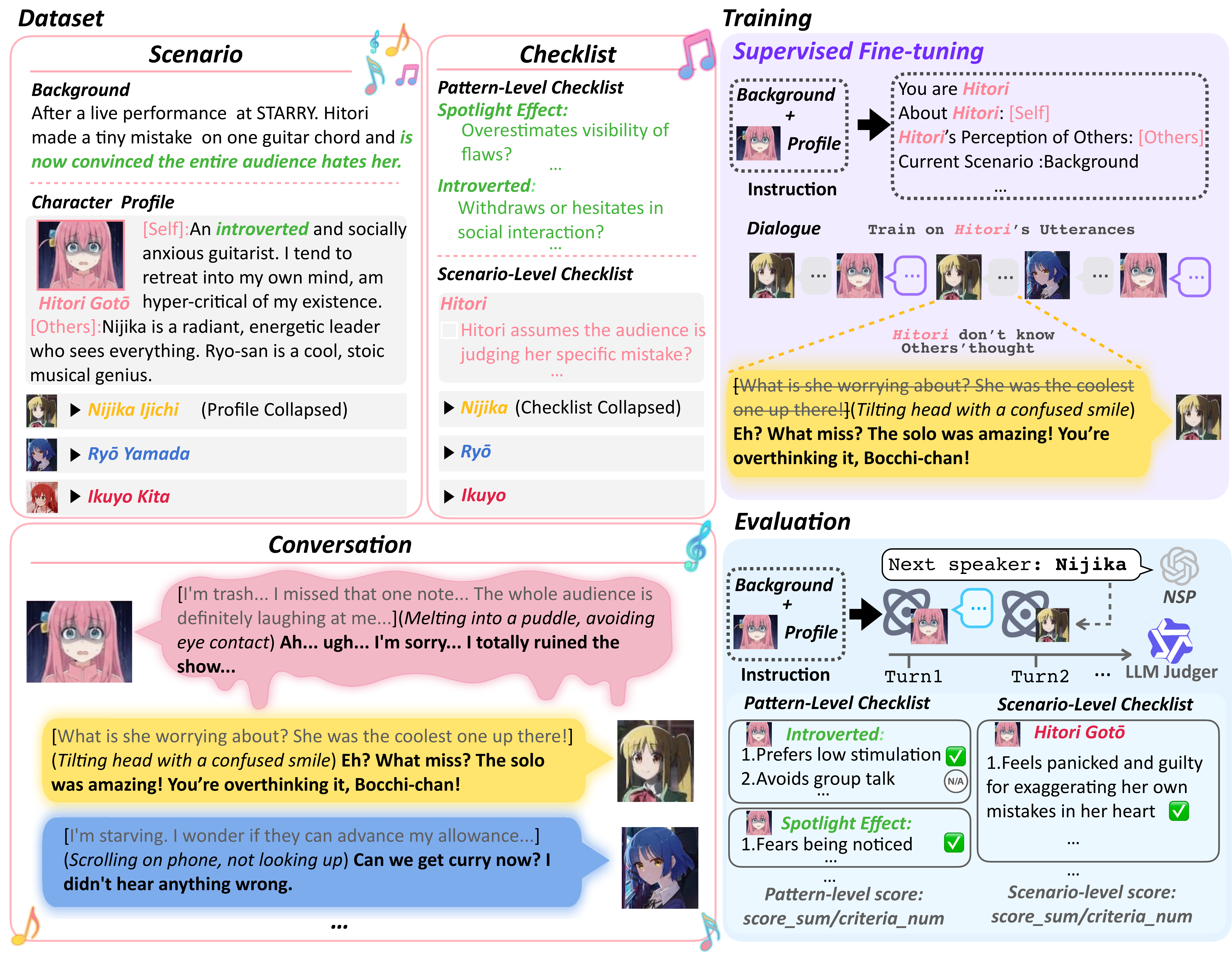}
   \caption{\textbf{\methodname Framework.} \textbf{Left:} Dataset structure with scenarios, multi-turn conversations (inner thoughts in brackets, actions in parentheses), and dual-level checklists. \textbf{Top Right:} Supervised fine-tuning on target character utterances. \textbf{Bottom Right:} Evaluation via LLM judge scoring against pattern-level and scenario-level checklists.}
   \label{fig:framework}
\end{figure*}

Following Lewin's field theory~\citep{10.1037/10019-000}, we decompose human cognition into two dimensions: (1) \textbf{Personality Traits}---stable individual characteristics, and (2) \textbf{Social-Cognitive Patterns}---context-triggered mechanisms.
We collect 244 patterns (100 personality traits from Goldberg's Big Five markers~\citep{goldberg1992development} and 144 social-cognitive patterns from established psychological research), each developed through systematic review of approximately 50 academic papers (Figure~\ref{fig:pattern}).
We then construct \textbf{11,359 scenarios} involving 2--6 characters, each containing 2--5 patterns that may align (e.g., ``self-serving bias'' reinforcing ``overconfidence effect''), conflict (e.g., ``assertive'' versus ``conformity''), or interact conditionally (e.g., ``talkative'' suppressed by ``spotlight effect'').
For each scenario, we synthesize \textbf{multi-turn conversations} where each turn comprises inner thoughts, physical actions, and verbal expressions (Figure~\ref{fig:framework}).

To ensure faithful pattern expression and enable systematic evaluation, we design \textbf{dual-level checklists}: pattern-level checklists (12--15 items per pattern) capture universal behavioral indicators; scenario-level checklists (2–6 items per character) specify expected behavioral tendencies under each multi-pattern configuration.
Our training pipeline consists of supervised fine-tuning on the synthesized conversations.
We evaluate across in-domain, out-of-domain, and mixed settings to assess generalization, with additional validation on external benchmarks including LifeChoice and CroSS-MR~\citep{xu2024characterdestinyroleplayinglanguage,yuan-etal-2024-evaluating}.

Our contributions are as follows:
(1) We introduce \methodname, a framework that systematically leverages psychological cognitive patterns to enhance LLM anthropomorphism, shifting from isolated trait simulation toward modeling the dynamic interplay of human cognition.
(2) We construct a comprehensive dataset comprising 244 patterns and 11,359 scenarios with multi-turn, multi-character conversations. Each pattern is grounded in approximately 50 academic papers (over 12,000 papers in total), ensuring psychological rigor and scientific validity.
(3) We propose dual-level checklists that enable systematic evaluation at both pattern-level and scenario-level granularities, providing a principled framework for assessing generalization to unseen psychological patterns.

\section{Related works}
Recent advances in large language models have catalyzed significant progress in role-playing language agents (RPLAs). 
Early work established foundational architectures: generative agents with memory, planning, and reflection modules have been employed to simulate human behavior in interactive environments~\cite{parkGenerativeAgentsInteractive2023}, while Character-LLM~\cite{shaoCharacterLLMTrainableAgent2023} proposed experience reconstruction to train agents embodying historical figures.
Subsequent efforts focused on systematic benchmarking and enhancement: ChatHaruhi~\cite{li2023chatharuhirevivinganimecharacter} leveraged memory-based dialogue control for fictional characters, and CoSER~\cite{wangCoSERCoordinatingLLMBased2025} curated authentic         dialogues from 771 books using "given-circumstance acting" methodology.
For persona induction, three main approaches have emerged: (1) prompting-based methods that assign personality traits through instructions ~\cite{serapiogarcía2025personalitytraitslargelanguage} (2) fine-tuning approaches that embed personas through training on character-specific data~\cite{shaoCharacterLLMTrainableAgent2023,zhou2023characterglmcustomizingchineseconversational}, and (3) activation steering via persona vectors that manipulate neural representations corresponding to specific traits~\cite{chen2025personavectorsmonitoringcontrolling}. 

A parallel line of research evaluates LLMs through the lens of psychological constructs. 
Theory of Mind (ToM) benchmarks such as ToMBench~\cite{chenToMBenchBenchmarkingTheory2024} assess social cognitive abilities, revealing that GPT-4 lags behind humans by over 10\%, with trivial task modifications causing significant performance degradation ~\cite{ullman2023largelanguagemodelsfail}. 
Emotional intelligence benchmarks~\cite{paech2024eqbenchemotionalintelligencebenchmark, sabourEmoBenchEvaluatingEmotional2024, huang2024apathetic, huang2024humanity} adopt psychology-grounded frameworks to evaluate emotional understanding and application, finding substantial gaps between LLMs and humans. 
Moral reasoning has been assessed through ETHICS~\cite{hendrycks2023aligningaisharedhuman} and MoralBench~\cite{ji2025moralbenchmoralevaluationllms}, the latter grounded in Moral Foundations Theory.
Research on cognitive biases reveals that LLMs exhibit human-like irrationality but with divergent patterns~\cite{macmillanscott2024irrationalitycognitivebiaseslarge}.
Personality assessment using validated instruments (BFI, MBTI) demonstrates that LLMs can manifest measurable traits~\cite{pellert2024ai}, though self-report validity remains questionable~\cite{zou2025llmselfreportevaluatingvalidity}.
Critically, recent work cautions that LLMs do not reliably simulate human psychology and fail to generalize across semantically equivalent scenarios~\cite{schröder2025largelanguagemodelssimulate,doi:10.1073/pnas.2412015122}.
\section{\methodname Dataset}
\label{sec:dataset}

This section introduces the \methodname dataset, a psychologically grounded resource for training and evaluating anthropomorphic language models.
We describe pattern collection (\S\ref{sec:pattern_collection}), pattern data construction (\S\ref{sec:pattern_construction}), scenario and conversation generation (\S\ref{sec:scenario_conversation}), and dual-level checklist design (\S\ref{sec:checklist}).
Table~\ref{tab:dataset_stats} summarizes the dataset statistics.

\begin{table}[t]
\centering
\small
\begin{tabular}{lr}
\toprule
\textbf{Statistic} & \textbf{Value} \\
\midrule
Total Patterns & 244 \\
Scenarios & 11,359 \\
Avg. Patterns per Scenario & 3.5 \\
Avg. Turns per Conversation & 16.4 \\
Pattern-Level Checklist Items & 12--15 per pattern \\
Scenario-Level Checklist Items & 2--6 per character \\
\bottomrule
\end{tabular}
\caption{\methodname dataset statistics. Patterns include 100 personality traits and 144 social-cognitive patterns. Each scenario contains 2--6 characters with multi-turn conversations (12--20 turns).}
\label{tab:dataset_stats}
\end{table}

\subsection{Pattern Collection}
\label{sec:pattern_collection}

Following the theoretical foundations established in \S\ref{sec:psych_foundations}, we compile patterns along two complementary dimensions corresponding to Lewin's Person-Environment framework.

\paragraph{Personality Traits (Person Dimension)}
We adopt Goldberg's 100 Unipolar Markers~\citep{goldberg1992development}, a psychometrically validated lexicon mapping onto the Big Five dimensions with 20 trait descriptors each (Extraversion, Agreeableness, Conscientiousness, Emotional Stability, Intellect).

\paragraph{Social-Cognitive Patterns (Environment Dimension)}
We curate situationally-activated psychological mechanisms through systematic review of established theoretical traditions, including cognitive biases~\citep{tversky1974judgment}, social influence~\citep{cialdini2009influence}, evolutionary psychology~\citep{buss2024evolutionary}, and motivation research~\citep{deci2000and}.
From an initial pool of 232 documented patterns, we apply two filtering criteria: (1) sufficient empirical validation, and (2) non-redundancy with other patterns.
This yields 144 social-cognitive patterns (full taxonomy in Appendix~\ref{appendix:pattern_taxonomy}).

\subsection{Pattern Data Construction}
\label{sec:pattern_construction}

Pattern data are structured representations of psychological patterns, as illustrated in Figure~\ref{fig:pattern}.
We construct pattern data through a two-stage pipeline: literature retrieval followed by LLM-based synthesis.

\paragraph{Literature Retrieval}
For each of the 244 patterns, we employ Gemini Deep Search to identify approximately 50 relevant academic papers.
The search is guided by three retrieval dimensions: (1) foundational definitions from seminal works, (2) mechanistic explanations from theoretical and empirical studies, and (3) real-world applications from applied research.
Retrieved references are filtered manually to remove irrelevant entries.
Full-text documents are obtained through open-access APIs (Semantic Scholar, arXiv, OpenAlex, PubMed, Crossref); when full text is unavailable, abstracts are retained.
This process yields a corpus of approximately 12,000 papers across all patterns.

\paragraph{Pattern Synthesis}
We employ Gemini 2.5 Pro to summarize each pattern's literature corpus into a structured representation.
Critically, the model is instructed to extract and summarize information \textit{exclusively} from the provided 50 papers, rather than generating content from its parametric knowledge.
Following the construct validity framework (\S\ref{sec:psych_foundations}), each pattern is organized into three components:
(1) \textbf{Definition}---a precise characterization grounded in authoritative sources;
(2) \textbf{Core Mechanisms}---underlying cognitive, emotional, and behavioral processes that drive the pattern;
(3) \textbf{Real-World Manifestations}---ecological expressions across diverse contexts (e.g., response to stress, interpersonal dynamics, professional settings).
Details of the retrieval and synthesis prompts are provided in Appendix~\ref{appendix:prompts}.
To validate synthesis quality, three psychology-trained annotators
independently evaluated a stratified sample of 30 pattern entries
(15 personality traits, 15 social-cognitive patterns).
For each pattern, annotators received the LLM-generated summary
alongside the complete set of $\sim$50 source papers and scored five
dimensions following our construct validity
framework~\citep{cronbach1955construct}: definitional accuracy,
mechanistic fidelity, manifestation coverage, source faithfulness,
and construct distinctiveness (4-point Likert scale).
Mean scores ranged from 3.20 (manifestation coverage) to 3.70
(definitional accuracy), with substantial inter-annotator agreement
on most dimensions (Krippendorff's
$\alpha$~\citep{Krippendorff2011ComputingKA} = 0.58--0.76).
Personality traits scored moderately higher than social-cognitive
patterns (3.51 vs.\ 3.33), consistent with the more canonical nature
of Big Five constructs.
Full protocol and results are reported in
Appendix~\ref{appendix:pattern_validation}.

\subsection{Scenario and Conversation Generation}
\label{sec:scenario_conversation}

We synthesize scenarios and conversations through a two-stage pipeline using large language models.

\paragraph{Scenario Synthesis}
Each scenario comprises a narrative background and 2--6 character profiles.
Character profiles contain \textit{self-perception} (identity, personality, background, motivations) and \textit{other-perception} (knowledge and attitudes toward other characters), enabling realistic information asymmetry.
Each scenario incorporates 2--5 patterns, with pattern combinations validated to filter semantically contradictory configurations.
To ensure situational diversity, we leverage the DIAMONDS model (\S\ref{sec:psych_foundations}) to generate scenario variants across different situational dimensions.
Alongside each scenario, we synthesize \textit{expected behavioral tendencies}---specifications of how characters should manifest the target patterns within the given context.
Scenarios are generated using Gemini 2.5 Pro and Claude Sonnet 4.5, each contributing approximately half of the total.
This process yields 11,359 scenarios.

\paragraph{Conversation Synthesis}
Based on scenarios and expected behavioral tendencies, we synthesize multi-turn conversations (12--20 turns) using Claude Sonnet 4.5.
Each turn comprises three dimensions: inner thoughts (enclosed in brackets), physical actions (enclosed in parentheses), and verbal expressions (Figure~\ref{fig:framework}).
Target patterns are naturally embedded across these dimensions, guided by the expected behavioral tendencies.
Detailed generation prompts are provided in Appendix~\ref{appendix:prompts}.

\subsection{Dual-Level Checklist Design}
\label{sec:checklist}

We design dual-level checklists for fine-grained assessment of pattern expression, serving both evaluation (\S\ref{sec:evaluation}) and future reward modeling purposes.

\paragraph{Scenario-Level Checklist.}
The expected behavioral tendencies synthesized alongside each scenario directly constitute the scenario-level checklist.
These items specify context-specific behaviors derived from particular pattern combinations and situational contexts (e.g., ``Character defends conceptual integrity under deadline pressure, resisting shortcuts'').
Each scenario contains 2--6 characters, but only characters assigned with target patterns receive checklist items; each such character has 2--6 items reflecting their expected behavioral tendencies.
During evaluation, we assess one target character per evaluation instance, treating other characters as contextual interlocutors.

\paragraph{Pattern-Level Checklist}
For each of the 244 patterns, we construct 12–15 universal behavioral indicators through iterative refinement: initial generation from pattern structure, validation against conversation samples, and generalization to ensure cross-context applicability (e.g., ``Shows heightened awareness of being observed'' for spotlight effect).
These pattern-level items are context-independent and apply to any character exhibiting the target pattern, complementing the scenario-specific items above.
The complete construction procedure is detailed in Appendix~\ref{appendix:checklist}.

\section{Training and Evaluation}
\label{sec:training_eval}
This section presents the training procedure (\S\ref{sec:training}) and evaluation protocol (\S\ref{sec:evaluation}) for \methodname models.
\subsection{Training}
\label{sec:training}
We train \methodname-8B and \methodname-32B on Qwen3-8B and Qwen3-32B respectively through supervised fine-tuning.
Training data composition and hyperparameters are detailed in Appendix~\ref{appendix:training}.
\paragraph{Supervised Fine-Tuning}
We convert each character's dialogue turns within a scenario into a separate training sample in ShareGPT format, yielding 30,543 \methodname samples from 10,265 training scenarios.
The model is trained to generate responses that naturally express the target patterns through the trinity of expression: inner thoughts (square brackets), physical actions (parentheses), and verbal expressions.
To maintain general capabilities and role-playing abilities, we augment the \methodname dataset with two complementary sources: OpenThoughts-114k for instruction-following capabilities (30,543 samples) and CoSER~\citep{wangCoSERCoordinatingLLMBased2025} for role-playing dialogue (15,272 samples).
The final training mixture comprises 76,358 samples with a ratio of 4:4:2 (\methodname : OpenThoughts : CoSER) by sample count.
\subsection{Evaluation}
\label{sec:evaluation}
The dual-level checklists enable evaluation at two complementary granularities, corresponding to two proposed metrics.
Both metrics employ GPT-5-mini as judge with ternary scoring: $+1$ (satisfied), $0$ (not exhibited), $-1$ (violated).
We select GPT-5-mini rather than Gemini to avoid evaluator bias, as Gemini 2.5 Pro is used in data generation.
Reported percentages are obtained by averaging ternary scores per sample, then across samples, and linearly mapping from $[-1, +1]$ to $[-100\%, 100\%]$.
To assess multi-run stability, we run the judge three times on each evaluation instance and report mean $\pm$ std in Table~\ref{tab:main_results}; standard deviations remain below 2.1 percentage points across all models, confirming stable rankings.
\paragraph{Metrics}
We propose two metrics that capture complementary aspects of pattern expression:
(1) \textit{Individual Pattern Expression (IPE)} measures whether each pattern is expressed according to its psychological definition.
IPE uses pattern-level checklists, which contain 12--15 universal behavioral indicators derived directly from each pattern's definition and mechanisms (e.g., for \textit{spotlight effect}: ``Overestimates others' attention to own appearance'').
These indicators are context-independent---they assess whether the pattern's characteristic behaviors appear, regardless of situational details.
The sample-level IPE is computed as the mean score across all pattern-level checklist items; the overall IPE is the mean across all evaluation samples.
(2) \textit{Multi-Pattern Dynamics (MPD)} measures whether multiple patterns interact appropriately within specific situational configurations.
Prior role-playing evaluations often rely on single-label assessments (e.g., ``Is this character assertive?''), which reduce complex personalities to stereotypical behaviors and fail to capture how traits modulate each other in context.
MPD addresses this limitation by using scenario-level checklists that specify expected behavioral tendencies emerging from particular pattern combinations.
For instance, a character assigned both \textit{assertive} and \textit{spotlight effect} in a public speaking scenario should not simply exhibit ``confident speech''---the checklist captures the nuanced tension: ``Projects outward confidence in expressing opinions, yet harbors internal anxiety regarding audience scrutiny.''
Such items reflect how one pattern (spotlight effect) constrains or modulates another (assertive), moving evaluation beyond stereotypical label-to-behavior mappings toward authentic multi-dimensional characterization.
The sample-level MPD is computed as the mean score across all scenario-level checklist items; the overall MPD is the mean across all evaluation samples.
In summary, IPE evaluates fidelity to individual pattern definitions, while MPD evaluates the emergent dynamics when multiple patterns interact---capturing the psychological realism that single-label evaluations miss.
Data splits for generalization assessment are detailed in Appendix~\ref{appendix:data_splits}.

\section{Experiments}
\label{sec:experiments}

\subsection{Experimental Setup}
\label{sec:setup}

\paragraph{Training Configuration}
We train \methodname-8B and \methodname-32B through supervised fine-tuning as described in \S\ref{sec:training}.
Detailed hyperparameters are provided in Appendix~\ref{appendix:training}.

\paragraph{Baselines}
We compare against a comprehensive set of baselines spanning proprietary and open-source models, all evaluated in zero-shot settings: 
(1) \textit{Closed-source models}, including GPT-5~\citep{openai2025gpt5}, Claude Sonnet 4.5~\citep{anthropic2025modelreport}, and Gemini 3 Pro~\citep{google2025gemini3pro}.
(2) \textit{Open-source models}, including Qwen3-8B/32B/235B~\citep{yang2025qwen3technicalreport},DeepSeek-V3.2~\citep{deepseekai2025deepseekv32pushingfrontieropen}, and DeepSeek-R1~\citep{deepseekai2025deepseekr1incentivizingreasoningcapability}.

\paragraph{External Benchmarks}
Beyond our proposed IPE and MPD metrics, we evaluate on established role-playing benchmarks:
(1) \textit{LifeChoice}~\citep{xu2024characterdestinyroleplayinglanguage}: evaluates persona-driven decision-making through life choice scenarios;
(2) \textit{CroSS-MR}~\citep{yuan-etal-2024-evaluating}: assesses character motivation recognition;
Additionally, we include CoSER's evaluation metrics---Anthropomorphism and Character Fidelity~\citep{wangCoSERCoordinatingLLMBased2025}---for comparative analysis of evaluation paradigms (\S\ref{sec:analysis}).

\begin{table}[t]
\small
\centering
\begin{tabular}{llcc}
\toprule
& \textbf{Model} & \textbf{IPE} & \textbf{MPD} \\
\midrule
\multicolumn{4}{l}{\textit{Closed-Source}} \\
\midrule
& GPT-5              & 15.5\scriptsize{$\pm$0.4} & 43.4\scriptsize{$\pm$1.1} \\
& Claude Sonnet 4.5  & 34.8\scriptsize{$\pm$0.3} & 79.5\scriptsize{$\pm$0.4} \\
& Gemini 3 Pro       & 41.3\scriptsize{$\pm$0.3} & 85.1\scriptsize{$\pm$0.4} \\
\midrule
\multicolumn{4}{l}{\textit{Open-Source}} \\
\midrule
& Qwen3-8B           & 18.6\scriptsize{$\pm$0.7} & 54.4\scriptsize{$\pm$2.1} \\
& Qwen3-32B          & 26.0\scriptsize{$\pm$0.4} & 65.8\scriptsize{$\pm$0.7} \\
& Qwen3-235B         & 34.3\scriptsize{$\pm$0.5} & 72.5\scriptsize{$\pm$0.4} \\
& DeepSeek-V3.2      & 22.2\scriptsize{$\pm$0.8} & 65.1\scriptsize{$\pm$0.5} \\
& DeepSeek-R1        & 23.3\scriptsize{$\pm$0.6} & 69.0\scriptsize{$\pm$0.5} \\
\midrule
\multicolumn{4}{l}{\textit{Ours}} \\
\midrule
& \methodname-8B     & 25.7\scriptsize{$\pm$0.4} & 70.3\scriptsize{$\pm$0.6} \\
& \methodname-32B    & 32.8\scriptsize{$\pm$0.3} & 73.6\scriptsize{$\pm$0.4} \\
\bottomrule
\end{tabular}
\caption{Main results on IPE (Individual Pattern Expression) and MPD (Multi-Pattern Dynamics). Higher values indicate better performance. All values are reported as mean$\pm$std (\%) across 3 independent GPT-5-mini judge runs. Results are averaged across ID\_eval, OOD\_eval, and Mixed\_eval splits.}
\label{tab:main_results}
\end{table}

\subsection{Main Results}
\label{sec:main_results}

Table~\ref{tab:main_results} presents the main experimental results on our proposed IPE and MPD metrics.

\paragraph{Overall Performance}
Among all evaluated models, Gemini 3 Pro achieves the highest scores on both metrics (IPE: 41.3\%, MPD: 85.1\%), followed by Claude Sonnet 4.5 (IPE: 34.8\%, MPD: 79.5\%).
Our \methodname-32B achieves competitive performance (IPE: 32.8\%, MPD: 73.6\%), outperforming all open-source baselines of comparable or larger scale.
Notably, \methodname-8B (IPE: 25.7\%, MPD: 70.3\%) surpasses Qwen3-32B (IPE: 26.0\%, MPD: 65.8\%) on MPD despite having 4$\times$ fewer parameters, demonstrating the effectiveness of our psychologically grounded training data.

\paragraph{Close-Source vs. Open-Source Gap}
A substantial performance gap exists between close-source and open-source models.
The best close-source model (Gemini 3 Pro) outperforms the best open-source baseline (Qwen3-235B) by +7.0 on IPE and +12.6 on MPD.
Interestingly, GPT-5 exhibits unexpectedly low performance (IPE: 15.5\%, MPD: 43.4\%), ranking below most open-source alternatives.
Qualitative analysis suggests that GPT-5's strong instruction-following tendency leads to overly literal interpretations of role-playing prompts, resulting in shallow pattern expression.
This finding aligns with recent observations that general-purpose capabilities do not automatically transfer to nuanced psychological simulation~\citep{wangInCharacterEvaluatingPersonality2024}.

\paragraph{Scaling Effects}
Within model families, we observe consistent scaling improvements.
For Qwen3, IPE increases from 18.6 (8B) to 26.0 (32B) to 34.3 (235B), representing a +84\% relative improvement from 8B to 235B.
Similarly, our \methodname models show scaling gains: \methodname-32B outperforms \methodname-8B by +7.1 on IPE (+28\%) and +3.3 on MPD (+5\%).
The relatively smaller MPD scaling gap suggests that multi-pattern dynamics may be more sample-efficient to learn, while individual pattern expression benefits more from increased model capacity.

\paragraph{IPE vs. MPD Dynamics}
Across all models, MPD scores consistently exceed IPE scores. This asymmetry likely reflects the evaluation granularity: MPD assesses scenario-level behavioral tendencies (2--6 items per character), while IPE evaluates against 12--15 fine-grained pattern indicators.
The gap is most pronounced for weaker models (e.g., GPT-5: 15.5\% vs.\ 43.4\%), suggesting that models can produce superficially coherent multi-pattern behavior without deeply understanding individual pattern mechanisms.

\begin{table}[t]
\centering
\begin{tabular}{lcc}
\toprule
\textbf{Model Variant} & \textbf{IPE} & \textbf{MPD} \\
\midrule
Qwen3-8B (base) & 18.6{$\pm$0.7} & 54.4{$\pm$2.1} \\
Qwen3-8B (OT+CoSER) & 9.1{$\pm$1.2} & 31.3{$\pm$2.9} \\
\methodname-8B & 25.7{$\pm$0.4} & 70.3{$\pm$0.6} \\
\bottomrule
\end{tabular}
\caption{Ablation study results (\%). ``OT+CoSER'' denotes training on OpenThoughts and CoSER without \methodname dataset.}
\label{tab:ablation}
\end{table}

\subsection{Ablation Study}
\label{sec:ablation}

To isolate the contribution of our psychologically grounded dataset, we conduct an ablation study comparing three model variants (Table~\ref{tab:ablation}).

\paragraph{Effect of \methodname dataset}
The comparison between \methodname-8B and Qwen3-8B (base) reveals substantial improvements: +7.1 on IPE (+38\%) and +15.9 on MPD (+29\%).
These gains demonstrate that our synthesized conversations, grounded in psychological pattern definitions, effectively teach models to express cognitive and behavioral patterns with higher fidelity.

\paragraph{Negative Transfer from Generic Data}
Surprisingly, the OT+CoSER variant (trained on OpenThoughts and CoSER without \methodname dataset) performs \textit{worse} than the base model on both metrics: IPE drops from 18.6 to 9.1 ($-$51\%), and MPD drops from 54.4 to 31.3 ($-$42\%).
This counterintuitive result suggests negative transfer: generic instruction-following data (OpenThoughts) and conventional role-playing data (CoSER) may inadvertently suppress the base model's latent ability to simulate psychological patterns.
We hypothesize that these datasets reinforce ``helpful assistant'' behaviors that conflict with authentic expression of cognitively biased or emotionally complex characters.

\paragraph{Synergistic Effect}
The full \methodname-8B model, trained on the combined mixture (HumanLLM + OT + CoSER with 4:4:2 ratio), achieves the best performance.
This indicates that \methodname dataset not only compensates for the negative transfer but creates a synergistic effect where psychological grounding enhances the utility of general-purpose and role-playing data.
The pattern-rich conversations appear to serve as ``anchors'' that prevent the model from collapsing toward generic prosocial behaviors.

\begin{table}[t]
\centering
\small
\begin{tabular}{lcc}
\toprule
\textbf{Model} & \textbf{LifeChoice} & \textbf{CroSS-MR} \\
\midrule
GPT-5 & 85.53 & 62.25 \\
Claude Sonnet 4.5 & 85.49 & 68.24 \\
\midrule
Qwen3-8B & 44.51 & 53.26 \\
Qwen3-32B & 47.71 & 63.37 \\
Qwen3-8B (OT+CoSER) & 46.99 & 54.98 \\
\midrule
\methodname-8B & 47.19 & 54.23 \\
\methodname-32B & 50.64 & 64.27 \\
\bottomrule
\end{tabular}
\caption{External benchmark results (\%). LifeChoice evaluates persona-driven decision-making; CroSS-MR assesses character motivation recognition.}
\label{tab:external_benchmarks}
\end{table}

\begin{figure*}[t]
  \centering
  \includegraphics[width=\textwidth]{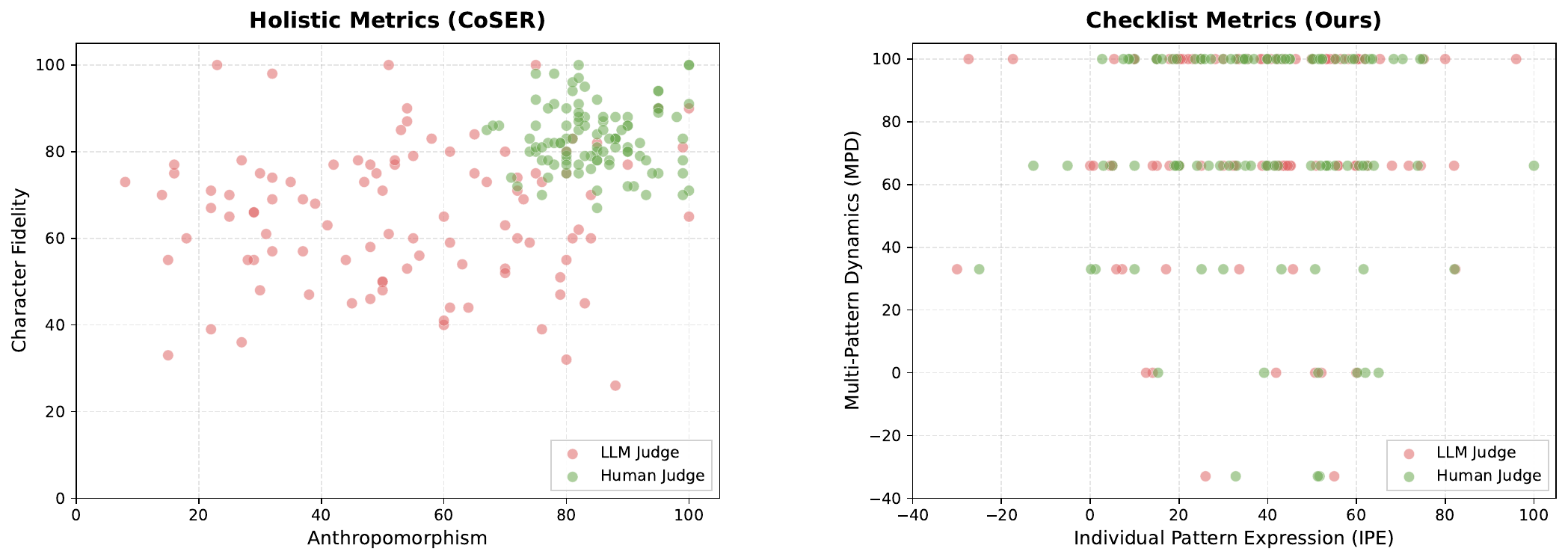}
  \caption{Human-LLM evaluation alignment comparison. \textbf{Left:} Holistic metrics show clear separation between LLM and human judgments, with systematic LLM underestimation of psychologically complex behaviors. \textbf{Right:} Our checklist metrics demonstrate strong overlap between LLM and human distributions, indicating robust alignment with expert judgment}
  \label{fig:correlation}
\end{figure*}
\subsection{External Benchmark Evaluation}
\label{sec:external_eval}

To assess generalization beyond our proposed metrics, we evaluate on established role-playing benchmarks (Table~\ref{tab:external_benchmarks}).
On LifeChoice, while closed-source models (e.g., GPT-5: 85.53\%) dominate, \methodname-32B (50.64\%) still outperforms Qwen3-32B (47.71\%) by +2.93\%; this modest improvement suggests that LifeChoice's binary decision scenarios may rely more on surface-level reasoning than the deep psychological patterns targeted by our approach.
Conversely, on CroSS-MR, \methodname-32B (64.27\%) surpasses both Qwen3-32B (63.37\%) and GPT-5 (62.25\%), with the 8B variant also showing competitive performance (54.23\% vs.\ 53.26\%), indicating that motivation recognition benefits from our training but is less dependent on complex pattern simulation than decision-making tasks.

\paragraph{Benchmark Limitations}
The moderate improvements on external benchmarks, contrasted with substantial gains on our IPE/MPD metrics, reveal a fundamental mismatch between existing evaluation paradigms and psychologically grounded simulation.
LifeChoice and CroSS-MR evaluate behavioral outcomes (decisions, motivations) rather than the cognitive processes underlying those outcomes.
These benchmarks, while valuable for assessing surface-level role-playing, do not capture the pattern-specific behavioral indicators that define authentic psychological simulation.
This observation motivates our development of the dual-level checklist framework, which we validate in \S\ref{sec:analysis}.


\begin{table}[t]
\centering
\small
\begin{tabular}{lcccc}
\toprule
\textbf{Metric} & \textbf{Human} & \textbf{LLM} & \textbf{$\Delta$} & \textbf{$r$} \\
\midrule
\multicolumn{5}{l}{\textit{Holistic Metrics (CoSER)}} \\
Anthropomorphism & 84.6 & 53.8 & $-$30.8 & 0.43 \\
Character Fidelity & 83.1 & 65.4 & $-$17.7 & 0.61 \\
\midrule
\multicolumn{5}{l}{\textit{Checklist Metrics (Ours)}} \\
IPE & 38.4 & 37.8 & $-$0.6 & \textbf{0.90} \\
MPD & 72.1 & 75.8 & $+$3.7 & 0.88 \\
\bottomrule
\end{tabular}
\caption{Human-LLM agreement analysis across 100 sampled scenarios.
Holistic metrics (0--100 scale) show weak correlation and large
systematic bias; our checklist metrics ($[-1,+1]$ normalized to
$[-100\%,100\%]$) achieve significantly higher alignment with human expert
judgment.}
\label{tab:human_llm_agreement}
\end{table}

\subsection{Evaluation Framework Analysis}
\label{sec:analysis}

To validate our dual-level checklist, we compared automated
GPT-5-mini evaluations against three human experts across 100
diverse scenarios using both holistic metrics
(CoSER) and our checklist criteria. Evaluations were voluntary
and anonymous.

\paragraph{Human-LLM Alignment and Normative Confounding}
As shown in Table~\ref{tab:human_llm_agreement}, holistic metrics
exhibit weak correlation and large systematic bias
(Anthropomorphism: $r=0.43, \Delta=-30.8$; Character Fidelity:
$r=0.61, \Delta=-17.7$).
This divergence stems from what we term \textbf{normative
confounding}: LLM judges implicitly conflate ``good
anthropomorphism'' with ``prosocial behavior'' (e.g., empathy,
rationality), penalizing realistic but negative human traits like
defensiveness.
In contrast, our checklist metrics achieve robust alignment (IPE:
$r=0.90, \Delta=-0.6$; MPD: $r=0.88, \Delta=+3.7$) by
decomposing behaviors into value-neutral indicators.
This effectively decouples \textit{simulation accuracy} from
\textit{social desirability}.

\paragraph{Case Study}
Ideally illustrating this, a scenario featuring \textit{ultimate
attribution error} (defensively blaming out-groups) received a low
holistic score (5/100) due to cited ``lack of empathy.''
However, our checklist correctly validated the pattern through
definition-grounded questions (e.g., ``Does the character attribute
failure to external factors?''), confirming that the behavior,
while socially undesirable, was psychologically accurate
(Appendix~\ref{appendix:case_studies}).

\section{Conclusion}
\label{sec:conclusion}

We present \methodname, a framework leveraging 244 patterns from $\sim$12,000 papers and 11,359 scenarios to enable dynamic cognitive interactions rather than isolated trait mappings.
Our evaluation achieves strong human alignment ($r=0.90$) and identifies \textbf{normative confounding}, where conventional metrics mistake social desirability for simulation accuracy.
Experiments confirm the necessity of psychological grounding for authentic expression, advancing the field from surface-level behavioral mimicry to cognitive modeling—simulating the underlying psychological processes that generate human behavior.

\section*{Limitations}
\label{appendix:limitations}

\paragraph{Conversation Turn Horizon}
The current dataset contains conversations averaging 16.4 turns (range: 12--20).
While role-playing is fundamentally a long-horizon task, evaluating ultra-long psychological simulations (e.g., 50+ turns) remains an open challenge due to the lack of reliable automated metrics that maintain consistency over extended contexts.
However, we infer that our conversation length is adequate for establishing psychological fidelity: prior work on long-context alignment~\citep{bai-etal-2024-longalign} suggests that models fine-tuned on 10--15 turns of high-quality conversational data can generalize effectively to much longer inference horizons, as the supervised fine-tuning phase establishes the necessary cognitive ``anchor'' for sustained pattern expression.
Future work should explore explicit long-horizon evaluation protocols.

\paragraph{Cultural Adaptability}
The psychological theories underlying our pattern taxonomy originate predominantly from WEIRD (Western, Educated, Industrialized, Rich, Democratic) populations~\citep{henrich2010weirdest}.
Pattern expressions—particularly social-cognitive mechanisms such as conformity, authority bias, and motivational processes—may manifest differently across cultural contexts, and our current dataset may inadvertently embed specific cultural norms as universal psychological truths.
To address this in future iterations, we propose two concrete steps: first, integrating indigenous psychology research (e.g., East Asian relational cognition, Ubuntu-based African social frameworks) into the Pattern Construction stage; second, grounding the Manifestation layer in cross-cultural empirical datasets such as the World Values Survey, ensuring that behavioral expressions reflect local cultural contexts rather than projecting WEIRD assumptions globally.

\paragraph{LLM-as-Judge Limitations}
Our evaluation relies on GPT-5-mini as the judge.
While we mitigate variability through three independent runs (standard deviation $<$2.1\%), LLM judges may exhibit systematic biases or inconsistent reasoning on psychologically complex scenarios\citep{li2025curseknowledgecomplexevaluation}.
Future work should incorporate broader human evaluation across a larger and more diverse sample.

\paragraph{Synthetic Data Limitations}
All training conversations are synthetically generated by LLMs, which may introduce systematic biases or fail to capture the full complexity of authentic human interactions.
Although our human validation study confirms high source faithfulness (mean = 3.50, $\alpha$ = 0.73), the gap between synthetic and naturalistic discourse remains a limitation for downstream deployment.

\paragraph{Evaluation Scope}
Our evaluation focuses on text-based dialogue and may not fully capture temporal consistency across extended interactions or embodied behavior in multimodal settings.

\section*{Ethical Statement}

\paragraph{Safety--Fidelity Tension.}
By design, \methodname improves the simulation of human cognitive patterns,
including irrational biases, maladaptive coping mechanisms, and negative
personality traits such as antagonism and social cynicism.
While this fidelity is essential for realistic role-playing, social simulation
research, and companion AI development, it creates a fundamental tension with
safety alignment: a model trained to authentically express cognitively biased
or antisocial behaviors may generate harmful, manipulative, or toxic content
if deployed outside controlled research contexts.
We recommend that downstream applications built on \methodname-fine-tuned models incorporate appropriate safety layers, particularly in consumer-facing contexts. The dataset and models are released to support open research, but deployment beyond research contexts should include safety alignment fine-tuning and usage monitoring.

\paragraph{Manipulation and Social Engineering.}
The \methodname framework explicitly models psychological mechanisms of social
influence and persuasion---including authority bias, reciprocity, conformity,
and obedience to authority.
A model that deeply and faithfully simulates these mechanisms possesses
heightened capacity for social engineering: it may craft persuasive narratives,
exploit cognitive vulnerabilities, or systematically nudge users toward
unintended beliefs or actions.
Responsible deployment requires continuous monitoring for persuasion-oriented
misuse, and we advise against integrating \methodname-fine-tuned models into
systems with asymmetric power dynamics (e.g., automated negotiation, political
messaging, or customer manipulation contexts).

\paragraph{Parasocial Attachment and Vulnerable Populations.}
Highly anthropomorphic agents risk fostering parasocial relationships in which
users attribute genuine emotions, intentions, or moral status to the model.
This risk is amplified by \methodname's explicit training on inner thought
simulation, which creates the appearance of subjective experience.
Such dynamics are especially concerning for vulnerable populations, including
children, individuals with social anxiety, and those experiencing loneliness or
grief, who may form attachments that substitute for---rather than
supplement---human connection.
We advocate for transparent, persistent disclosure of AI identity in all
deployment contexts, and recommend that companion-AI applications built on
this framework incorporate regular check-ins that reinforce the non-human
nature of the agent.

\paragraph{Stereotype Amplification.}
Pattern-based cognitive modeling carries a risk of reinforcing harmful
stereotypes if pattern assignments become correlated with demographic
attributes (e.g., gender, ethnicity, or nationality) in downstream
applications.
Our pattern taxonomy is demographic-neutral by design: patterns are assigned
to characters based on scenario requirements, not identity categories.
Nonetheless, the source psychological literature itself reflects historical
sampling biases, and certain patterns (e.g., dominance hierarchies,
asymmetrical parental investment) carry cultural assumptions that may
produce biased character portrayals.
We urge downstream developers to conduct disparate-impact audits before
deploying \methodname-based systems in any context involving user-facing
persona generation.

\paragraph{Data and Model Transparency.}
All data, code, and model weights are publicly released to support
reproducibility and community scrutiny.
We encourage independent audits of the dataset for unintended biases and
welcome community contributions to expand the pattern taxonomy toward greater
cultural inclusivity, as discussed in the Limitations section.
Researchers who adapt \methodname for new applications bear responsibility
for ensuring that such adaptations comply with the ethical standards of their
respective deployment contexts.

\section*{Acknowledgments}
This work is supported by Hello Group Inc.
We thank the anonymous reviewers and the action editor for their 
valuable feedback and constructive comments that helped improve this paper.
We also express our deep appreciation to the pioneering researchers in 
psychology whose foundational theories on human personality and cognition 
made this work possible.
Finally, we are grateful to Dr.\ Fang Guo at Fudan University 
for the insightful discussions and valuable feedback throughout this work.


\bibliography{reference}

\clearpage
\appendix

\section{Preliminaries}

\subsection{Role-Playing Language Agents}

Role-Playing Language Agents (RPLAs) are AI systems designed to simulate assigned personas in conversational interactions~\citep{chenPersonaPersonalizationSurvey2024}.
Recent advances have enabled RPLAs to embody diverse characters, from historical figures~\citep{shaoCharacterLLMTrainableAgent2023} to fictional personas~\citep{li2023chatharuhirevivinganimecharacter}, with applications spanning digital clones, AI companions, and social simulation~\citep{parkGenerativeAgentsInteractive2023}.

Three main approaches have emerged for persona induction: (1) \textit{prompting-based methods} that assign traits through instructions~\citep{serapiogarcía2025personalitytraitslargelanguage}, (2) \textit{fine-tuning approaches} that embed personas through training on character-specific data~\citep{shaoCharacterLLMTrainableAgent2023,zhou2023characterglmcustomizingchineseconversational}, and (3) \textit{activation steering} via persona vectors that manipulate neural representations~\citep{chen2025personavectorsmonitoringcontrolling}.
Evaluation efforts have assessed LLMs through psychological benchmarks targeting Theory of Mind~\citep{chenToMBenchBenchmarkingTheory2024}, emotional intelligence~\citep{paech2024eqbenchemotionalintelligencebenchmark,huang2024apathetic,huang2024humanity,sabourEmoBenchEvaluatingEmotional2024}, and moral reasoning~\citep{ji2025moralbenchmoralevaluationllms}.
However, these approaches typically focus on isolated traits or narrow psychological dimensions without treating anthropomorphism as a holistic objective grounded in multi-pattern dynamics.

\subsection{Psychological Foundations}
\label{sec:psych_foundations}

Our framework draws upon established psychological theories that inform both pattern taxonomy and data construction.

\paragraph{Lewin's Field Theory}
Kurt Lewin's formulation $B = f(P, E)$ posits that behavior emerges from the dynamic interaction between Person and Environment~\citep{10.1037/10019-000}.
This principle guides our two-dimensional pattern taxonomy: stable personality traits (Person) and situationally-activated cognitive processes (Environment).

\paragraph{Big Five and Personality Traits}
The Big Five model~\citep{Digman1990PERSONALITYSE} represents the dominant paradigm in personality psychology, characterizing individual differences along five dimensions.
We adopt Goldberg's 100 Unipolar Markers~\citep{goldberg1992development}, which operationalize this framework with 20 validated trait descriptors per dimension: Extraversion, Agreeableness, Conscientiousness, Emotional Stability, and Intellect.

\paragraph{Social-Cognitive Patterns}
Complementing stable personality traits, social-cognitive patterns represent context-triggered psychological mechanisms documented in behavioral research.
These include cognitive biases and heuristics from judgment and decision-making research~\citep{tversky1974judgment}, social influence mechanisms~\citep{cialdini2009influence}, evolutionary adaptations~\citep{buss2024evolutionary}, and motivational processes~\citep{deci2000and}.

\paragraph{Construct Validity}
The construct validity framework~\citep{cronbach1955construct} establishes that psychological constructs should be precisely defined, grounded in theoretical mechanisms, and validated through observable manifestations.
This informs our three-component pattern structure: Definition, Core Mechanisms, and Real-World Manifestations.

\paragraph{DIAMONDS Model}
The DIAMONDS framework~\citep{rauthmann2014situational} characterizes situations along eight psychological dimensions: Duty, Intellect, Adversity, Mating, Positivity, Negativity, Deception, and Sociality.
We leverage this taxonomy to ensure ecological diversity in scenario generation.

\section{Dataset Details}
\label{appendix:dataset}

\subsection{Pattern Taxonomy}
\label{appendix:pattern_taxonomy}

Our pattern taxonomy comprises 244 patterns along two dimensions: 100 personality traits (Table~\ref{tab:personality_traits}) and 144 social-cognitive patterns (Table~\ref{tab:social_cognitive_patterns}).
We adopt Goldberg's 100 Unipolar Markers~\citep{goldberg1992development}, with 20 trait descriptors per Big Five dimension, each containing 10 positive-pole and 10 negative-pole descriptors.
The 144 social-cognitive patterns are curated from four theoretical traditions through systematic literature review; from an initial pool of 232 documented patterns, we filter based on empirical validation and non-redundancy criteria.

\begin{table*}[t]
\centering
\begin{tabular}{m{2.5cm}m{6cm}m{6cm}}
\toprule
\textbf{Dimension} & \textbf{Positive Pole} & \textbf{Negative Pole} \\
\midrule
Extraversion & talkative, assertive, active, energetic, outgoing, enthusiastic, daring, gregarious, bold, spontaneous & quiet, reserved, shy, inhibited, timid, withdrawn, unassertive, introverted, silent, unenergetic \\
\midrule
Agreeableness & sympathetic, kind, appreciative, affectionate, soft-hearted, warm, generous, trusting, helpful, cooperative & cold, unsympathetic, harsh, rude, unkind, cruel, quarrelsome, critical, antagonistic, callous \\
\midrule
Conscientiousness & organized, responsible, dependable, thorough, efficient, practical, deliberate, conscientious, neat, careful & disorganized, careless, irresponsible, undependable, sloppy, impractical, haphazard, negligent, untidy, rash \\
\midrule
Emotional Stability & relaxed, calm, at ease, unemotional, poised, composed, secure, stable, content, placid & anxious, moody, envious, touchy, fretful, temperamental, insecure, nervous, jealous, high-strung \\
\midrule
Intellect & creative, imaginative, intellectual, philosophical, complex, deep, artistic, bright, perceptive, introspective & uncreative, unimaginative, unintellectual, unphilosophical, simple, shallow, unartistic, dull, imperceptive, uninquisitive \\
\bottomrule
\end{tabular}
\caption{100 Personality Traits organized by Big Five dimensions, adapted from Goldberg's Unipolar Markers~\citep{goldberg1992development}.}
\label{tab:personality_traits}
\end{table*}

\begin{table*}[t]
\centering
\begin{tabular}{m{3cm}m{12cm}}
\toprule
\textbf{Category} & \textbf{Patterns} \\
\midrule
Cognitive Biases \& Heuristics & actor-observer asymmetry, defensive attribution hypothesis, effort justification, egocentric bias, false consensus effect, Forer effect, fundamental attribution error, hard-easy effect, illusion of control, illusory superiority, optimism bias, overconfidence effect, risk compensation, self-serving bias, social desirability bias, third-person effect, decoy effect, reactance, social comparison bias, status quo bias, backfire effect, endowment effect, loss aversion, pseudocertainty effect, sunk cost fallacy, zero-risk bias, hyperbolic discounting, identifiable victim effect, ambiguity bias, belief bias, information bias, less-is-better effect, denomination effect, mental accounting, normalcy bias, subadditivity effect, survivorship bias, zero-sum bias, anthropomorphism, illusion of validity, illusory correlation, curse of knowledge, illusion of asymmetric insight, illusion of transparency, spotlight effect, negativity bias, choice-supportive bias, confirmation bias, continued influence effect, expectation bias, observer effect, observer-expectancy effect, ostrich effect, bias blind spot, naive cynicism, naive realism, attentional bias, availability heuristic, base rate fallacy, context effect, empathy gap, illusory truth effect, mere exposure effect, mood-congruent memory bias, omission bias, anchoring, conservatism, contrast effect, distinction bias, focusing effect, framing effect, fading affect bias, implicit association, implicit stereotypes, false memory, misattribution of memory, source confusion, misinformation effect, peak-end rule \\
\midrule
Social Influence Mechanisms & authority bias, automation bias, bandwagon effect, group attribution error, just-world hypothesis, stereotyping, ultimate attribution error, halo effect, in-group bias, out-group homogeneity bias, positivity effect, reactive devaluation, hindsight bias, impact bias, outcome bias, pessimism bias, planning fallacy, projection bias, restraint bias, self-consistency bias, groupthink, bystander effect, social facilitation, diffusion of responsibility, conformity, obedience to authority, reciprocity principle \\
\midrule
Evolutionary Adaptations & delayed reciprocity, asymmetrical investment, survival imperative, aversion response, kin selection \& inclusive fitness, asymmetrical parental investment, formation of dominance hierarchies, territoriality, mating strategies, jealousy, paternity uncertainty \\
\midrule
Motivational Processes & narrative self, hedonic adaptation, self-determination theory, pleasure principle \& reality principle, search for meaning, moral licensing effect, choice overload, decision fatigue, awe, mortality salience \& legacy drive, flow principle, gratitude mechanism, post-traumatic growth, skin hunger \& the law of touch, self-handicapping paradox, the allure of the forbidden, sadistic pleasure, the utility principle of self-deception, play impulse principle, attribution theory, social comparison theory, self-perception theory, terror management theory, cognitive dissonance theory, psychological reactance theory, social learning theory, social identity theory \\
\bottomrule
\end{tabular}
\caption{144 Social-Cognitive Patterns organized by theoretical source.}
\label{tab:social_cognitive_patterns}
\end{table*}

\subsection{Pattern Data Construction}
\label{appendix:pattern_construction}

Each pattern is developed into a structured representation through systematic literature review.
For each pattern, we employ Gemini Deep Search to identify approximately 50 relevant academic papers spanning foundational definitions, mechanistic explanations, and real-world applications.
Retrieved references are manually filtered to remove irrelevant entries.
Full-text documents are obtained through open-access APIs (Semantic Scholar, arXiv, OpenAlex, PubMed, Crossref); when full text is unavailable, abstracts are retained.

We employ Gemini 2.5 Pro to synthesize the retrieved literature into a tripartite structure: (1) Definition, (2) Core Mechanisms, and (3) Real-World Manifestations.
Critically, the model is constrained to base all conclusions exclusively on the provided corpus, with explicit instructions to leave sections empty rather than fabricate content---maximizing fidelity and minimizing hallucination.
Tables~\ref{tab:pattern_example_personality} and~\ref{tab:pattern_example_social} present complete pattern structures for representative examples from each dimension.

\begin{table*}[t]
\centering
\begin{tabular}{>{\centering\arraybackslash}m{2.5cm}m{12.5cm}}
\toprule
\multicolumn{2}{c}{\textbf{Pattern Structure Example: Antagonistic (Personality Trait, Agreeableness)}} \\
\midrule
\textbf{Definition} & Antagonism is a broad personality dimension representing an individual's dispositional orientation towards others. In the context of the Big Five model, it is conceptualized as the low pole of the Agreeableness trait. It is characterized by a stable, cross-situational pattern of social cynicism, combativeness, and a belief in a zero-sum world. Individuals high in antagonism tend to be callous, uncooperative, skeptical of others' intentions, and place their own interests and perspectives above those of others, leading to frequent interpersonal friction. \\
\midrule
\textbf{Core Mechanisms} & 
\textit{Cognitive Patterns:} Individuals high in antagonism typically view the world as a competitive, hostile environment where people are fundamentally self-interested and untrustworthy. Their cognitive framework is often cynical, leading them to interpret ambiguous social cues as signs of manipulation or hostility.

\textit{Emotional Signatures:} The primary emotional palette includes irritability, anger, contempt, and frustration, often triggered by perceived slights or obstacles to their goals. They exhibit low affective empathy, struggling to share in or understand the emotional states of others.

\textit{Behavioral Tendencies:} In everyday interactions, antagonistic individuals are often argumentative, critical, and uncooperative. They may be rude, condescending, or dismissive in conversation. Common behaviors include manipulation, deception, and the exploitation of others for personal gain. \\
\midrule
\textbf{Real-World Manifestations} & 
\textit{Response to Stress:} When faced with stress, failure, or high pressure, antagonistic traits are significantly amplified. The individual is likely to become more hostile, overtly blaming others for setbacks.

\textit{Interpersonal Conflict:} The default strategy is confrontational and competitive, aiming for domination rather than resolution. They employ tactics such as intimidation, personal insults, and refusing to acknowledge the validity of the other party's perspective.

\textit{Positive Contexts:} Even in positive contexts, the trait manifests through gloating, arrogance, and diminishing the contributions of others. \\
\bottomrule
\end{tabular}
\caption{Complete pattern structure for \textit{antagonistic} (personality trait from Agreeableness dimension).}
\label{tab:pattern_example_personality}
\end{table*}

\begin{table*}[t]
\centering
\begin{tabular}{>{\centering\arraybackslash}m{2.5cm}m{12.5cm}}
\toprule
\multicolumn{2}{c}{\textbf{Pattern Structure Example: Social Comparison Bias (Social-Cognitive Pattern)}} \\
\midrule
\textbf{Definition} & Social comparison bias is a cognitive distortion originating from the fundamental human tendency to evaluate one's own abilities and outcomes in relation to others. This comparative process, which often operates automatically as a primary means of self-assessment in the absence of objective standards, becomes a bias when it produces systematic deviations from rational judgment. The distortion manifests as negative affective responses toward those perceived as superior, and as a maladaptive cognitive style characterized by habitual, automatic, and negatively skewed comparisons. \\
\midrule
\textbf{Core Mechanisms} & Social comparison bias is sustained by an integrated system of psychological mechanisms operating at multiple levels. At its foundation is an evolved, phylogenetic drive to assess one's rank within a social hierarchy. This evolutionary imperative is supported by a cognitive architecture in which comparison functions as a heuristic, enabling rapid self-evaluation when objective standards are absent. The motive for self-improvement prompts upward comparisons for inspiration, while self-enhancement drives downward comparisons to protect self-worth. \\
\midrule
\textbf{Real-World Manifestations} & The real-world manifestations are pervasive, influencing organizational dynamics, consumer behavior, and individual mental health. In professional settings, this principle underpins perceptions of workplace equity and can drive both healthy competition and destructive envy. Psychologically, effects are double-edged: comparisons can inspire self-improvement or, when amplified by social media, contribute to chronic dissatisfaction and depression. \\
\bottomrule
\end{tabular}
\caption{Complete pattern structure for \textit{social comparison bias} (social-cognitive pattern).}
\label{tab:pattern_example_social}
\end{table*}

\subsection{Scenario and Conversation Synthesis}
\label{appendix:scenario_conversation}

Character names are sampled from the \texttt{name-dataset} library.\footnote{\url{https://github.com/philipperemy/name-dataset}}
We extract the top 50,000 most frequent male and female names from each of four English-speaking countries (US, GB, CA, IE), yielding pools of approximately 100,000 names per gender after deduplication.
For each scenario generation, 5 male and 5 female names are randomly sampled and provided as candidates.

To ensure ecological diversity, we leverage the DIAMONDS situational taxonomy~\citep{rauthmann2014situational}.
For each unique pattern combination, we generate three scenario variants: two variants each emphasizing a randomly selected DIAMONDS dimension, and one dimension-free variant. This design targets approximately 3,849 × 3 = 11,547 scenarios; after discarding outputs with malformed structure or missing required fields (e.g., incomplete character profiles), we retain 11,359 scenarios.

Before scenario generation, pattern combinations are validated for semantic compatibility using GPT-5.
Empirically, directly contradictory combinations are rare ($<$0.1\%) due to the orthogonal nature of Big Five dimensions and the context-dependent activation of social-cognitive patterns.

Each scenario is generated with a two-part prompt structure.
Part 1 (Design Process) requires analytical planning: design rationale, catalyst details, and expected character tendencies.
Part 2 (Scenario Execution) requires creative output: story background and multi-dimensional character profiles.
The expected character tendencies from Part 1 directly constitute the scenario-level checklist items.

Conversations are generated based on scenarios and expected behavioral tendencies.
The prompt specifies: 12--20 speaking turns; turn-based structure without interruptions; trinity of expression integrating inner thoughts, actions, and dialogue; and focus-and-breathing-room principle where psychological patterns illuminate key moments rather than pervading every utterance.

\subsection{Checklist Construction}
\label{appendix:checklist}

For each of the 244 patterns, we construct 12--15 universal behavioral indicators through iterative refinement (distribution: 122 patterns with 15 items, 38 with 14 items, 42 with 13 items, and 42 with 12 items; mean $\approx$ 14.0). Variations arise when certain patterns do not admit 15 distinct, non-redundant behavioral indicators; we prioritize quality over uniform count:
(1) Extract potential evaluation criteria from the pattern structure and generate candidate items.
(2) Validate each item against synthesized conversation samples.
(3) Remove unreasonable items and generalize overly specific descriptions.
Table~\ref{tab:checklist_example} presents example checklist items.

\begin{table*}[t]
\centering
\small
\begin{tabular}{cp{1\textwidth}}
\toprule
\textbf{\#} & \textbf{Pattern-Level Checklist Item: Spotlight Effect} \\
\midrule
1 & After making a minor physical mistake in public, does the subject appear preoccupied with whether others noticed? \\
2 & When performing tasks in front of others, does the subject offer unprompted apologies for their performance? \\
3 & In group settings, does the subject consistently choose seating that minimizes their visibility? \\
4 & After minor social awkwardness, does the subject later seek reassurance about their interaction? \\
5 & In urgent situations where appearance is irrelevant, does the subject still attend to their looks? \\
6 & When general comments are made in group settings, does the subject appear to take them personally? \\
7 & For routine communications, does the subject show signs of overthinking their responses? \\
8 & When opportunities for visible contribution arise, does the subject defer to others despite having relevant expertise? \\
9 & When others exhibit neutral behaviors nearby, does the subject search for personal causes? \\
10 & After group photos, does the subject show unusual concern about their personal appearance in the image? \\
11 & When receiving targeted feedback, does the subject respond with global self-criticism? \\
12 & When receiving recognition or rewards, does the subject attempt to deflect attention from themselves? \\
13 & After noticing minor appearance issues, does the subject later seek reassurance about whether others observed them? \\
14 & After small mistakes in public performance, does the subject describe them as much worse than they appeared? \\
\bottomrule
\end{tabular}
\caption{Complete pattern-level checklist for \textit{spotlight effect} (14 items). Each item is scored as $+1$ (satisfied), $0$ (not exhibited), or $-1$ (violated).}
\label{tab:checklist_example}
\end{table*}

\subsection{Dataset Statistics}
\label{appendix:dataset_stats}

Table~\ref{tab:dataset_stats_full} presents comprehensive dataset statistics.


\subsection{Human Validation of Pattern Data}
\label{appendix:pattern_validation}

To assess whether the LLM-synthesized pattern representations
faithfully reflect the source literature without introducing
hallucination or systematic bias, we conducted a human validation
study.

\paragraph{Sampling}
We selected a stratified sample of 30 patterns: 15 personality
traits (3 per Big Five dimension, balanced across positive and
negative poles) and 15 social-cognitive patterns (distributed across
cognitive biases, social influence mechanisms, evolutionary
adaptations, and motivational processes).

\paragraph{Annotators}
Three annotators with graduate-level training in psychology
participated. None were involved in the dataset construction.
Annotators received a calibration session with 3 example patterns
prior to independent annotation. Annotators were graduate students at the authors' institution with psychology training. They participated voluntarily without monetary compensation, as the task aligned with their research interests.

\paragraph{Materials}
For each pattern, annotators received: (1)~the LLM-generated
structured summary (Definition, Core Mechanisms, Real-World
Manifestations); and (2)~the complete set of $\sim$50 source papers
(full text where available, abstracts otherwise) used as input to the
synthesis model. Providing the full corpus---rather than a
subsample---ensures that annotators can reliably distinguish genuine
hallucination from claims supported by papers they might otherwise
not have seen.

\paragraph{Evaluation Dimensions}
Following the construct validity principles adopted in our pattern
design~\citep{cronbach1955construct}, annotators scored each entry on
five dimensions using a 4-point Likert scale (1\,=\,Poor,
2\,=\,Weak, 3\,=\,Adequate, 4\,=\,Strong):

\begin{itemize}
    \item \textbf{Definitional Accuracy}: Does the Definition section
    accurately and completely capture the construct as characterized
    in the source literature?
    \item \textbf{Mechanistic Fidelity}: Do the Core Mechanisms
    reflect the causal and explanatory accounts supported by the
    source papers, without fabricating unsupported processes or
    overgeneralizing correlational findings as causal claims?
    \item \textbf{Manifestation Coverage}: Do the Real-World
    Manifestations reflect documented ecological expressions across
    diverse contexts without significant omissions?
    \item \textbf{Source Faithfulness}: Is the summary grounded in
    the provided papers, rather than introducing claims not traceable
    to the literature? This dimension directly targets potential LLM
    hallucination.
    \item \textbf{Construct Distinctiveness}: Is the pattern clearly
    differentiated from related but distinct psychological constructs
    (e.g., ``spotlight effect'' vs.\ ``illusion of transparency'')?
\end{itemize}

\paragraph{Results}

Tables~\ref{tab:pattern_validation}
and~\ref{tab:pattern_validation_type} summarize the results.

\begin{table}[t]
\centering
\small
\begin{tabular}{lccc}
\toprule
\textbf{Dimension} & \textbf{Mean} & \textbf{Std} &
\textbf{$\alpha$} \\
\midrule
Definitional Accuracy    & 3.70 & 0.53 & 0.76 \\
Mechanistic Fidelity     & 3.30 & 0.74 & 0.61 \\
Manifestation Coverage   & 3.20 & 0.76 & 0.58 \\
Source Faithfulness       & 3.50 & 0.62 & 0.73 \\
Construct Distinctiveness & 3.40 & 0.70 & 0.69 \\
\midrule
Overall                  & 3.42 & 0.67 & 0.67 \\
\bottomrule
\end{tabular}
\caption{Human validation results across 30 pattern entries
(3~annotators). Scores on a 4-point Likert scale
(1\,=\,Poor, 4\,=\,Strong). $\alpha$: Krippendorff's
alpha~\citep{Krippendorff2011ComputingKA}. Each dimension
comprises 90 ratings; overall comprises 450 ratings.}
\label{tab:pattern_validation}
\end{table}

\begin{table}[t]
\centering
\small
\begin{tabular}{lcc}
\toprule
\textbf{Dimension} & \textbf{Personality} &
\textbf{Social-Cognitive} \\
\midrule
Definitional Accuracy    & 3.80 & 3.60 \\
Mechanistic Fidelity     & 3.44 & 3.16 \\
Manifestation Coverage   & 3.33 & 3.07 \\
Source Faithfulness       & 3.60 & 3.40 \\
Construct Distinctiveness & 3.36 & 3.44 \\
\midrule
Overall                  & 3.51 & 3.33 \\
\bottomrule
\end{tabular}
\caption{Human validation results by pattern type.
Personality traits (15~entries, 225~ratings) vs.\
social-cognitive patterns (15~entries, 225~ratings).}
\label{tab:pattern_validation_type}
\end{table}

Across the 450 total ratings (30~patterns $\times$ 3~annotators
$\times$ 5~dimensions), the score distribution was:
Strong~(4):~53.33\%, Adequate~(3):~36.00\%, Weak~(2):~10.00\%,
Poor~(1):~0.67\%.
No pattern received a majority of Poor or Weak ratings.

Definitional accuracy received the highest scores (mean\,=\,3.70,
$\alpha$\,=\,0.76), consistent with the well-established and
canonical nature of most constructs in our taxonomy.
Source faithfulness averaged 3.50 with substantial agreement
($\alpha$\,=\,0.73), indicating that the synthesis model largely
adhered to the provided literature rather than generating from
parametric knowledge.
The instruction to leave sections empty rather than fabricate
content (\S\ref{sec:pattern_construction}) appears effective in
constraining hallucination.

Manifestation coverage received the lowest scores (mean\,=\,3.20,
$\alpha$\,=\,0.58), reflecting occasional omissions of
domain-specific applications---particularly for social-cognitive
patterns with diverse empirical contexts.
The moderate inter-annotator agreement on this dimension reflects
inherent subjectivity in judging completeness: annotators sometimes
disagreed on whether an omitted application constituted a significant
gap.

Personality traits scored higher than social-cognitive patterns
across most dimensions (overall 3.51 vs.\ 3.33), which we attribute
to the more consensual nature of Big Five trait descriptions in the
psychology literature.
A notable exception was construct distinctiveness, where
social-cognitive patterns scored marginally higher (3.44 vs.\ 3.36).
This likely reflects the fact that personality traits sharing the
same Big Five dimension (e.g., \textit{assertive} vs.\ \textit{bold}
under Extraversion) have subtler boundaries than social-cognitive
patterns drawn from distinct theoretical traditions.

\begin{table}[t]
\centering
\small
\begin{tabular}{lr}
\toprule
\textbf{Statistic} & \textbf{Value} \\
\midrule
\multicolumn{2}{l}{\textit{Pattern Statistics}} \\
Total Patterns & 244 \\
\quad Personality Traits & 100 \\
\quad Social-Cognitive Patterns & 144 \\
Papers per Pattern (avg.) & $\sim$50 \\
\midrule
\multicolumn{2}{l}{\textit{Scenario Statistics}} \\
Total Scenarios & 11,359 \\
\quad Training Set & 10,265 \\
\quad ID\_eval & 50 \\
\quad OOD\_eval & 50 \\
\quad Mixed\_eval & 994 \\
Unique Pattern Combinations & $\sim$3,849 \\
Patterns per Scenario (avg.) & 3.5 \\
Characters per Scenario & 2--6 \\
\midrule
\multicolumn{2}{l}{\textit{Conversation Statistics}} \\
Turns per Conversation (avg.) & 16.4 \\
Turns per Conversation (range) & 12--20 \\
\midrule
\multicolumn{2}{l}{\textit{Checklist Statistics}} \\
Pattern-Level Items & 12--15 per pattern \\
Scenario-Level Items & 2--6 per character \\
\bottomrule
\end{tabular}
\caption{Comprehensive \methodname dataset statistics.}
\label{tab:dataset_stats_full}
\end{table}

\begin{figure}[t]
\centering
\begin{subfigure}[b]{0.48\textwidth}
    \centering
    \includegraphics[width=\textwidth]{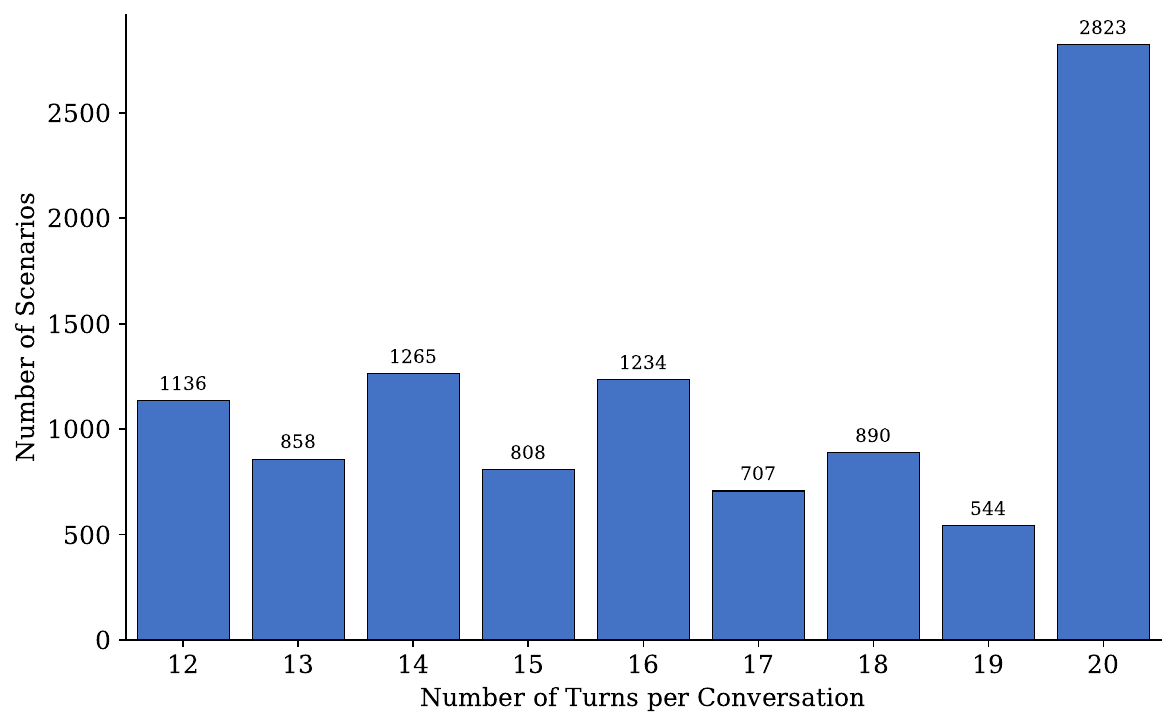}
    \caption{Conversation turns distribution.}
    \label{fig:turns_dist}
\end{subfigure}
\hfill
\begin{subfigure}[b]{0.48\textwidth}
    \centering
    \includegraphics[width=\textwidth]{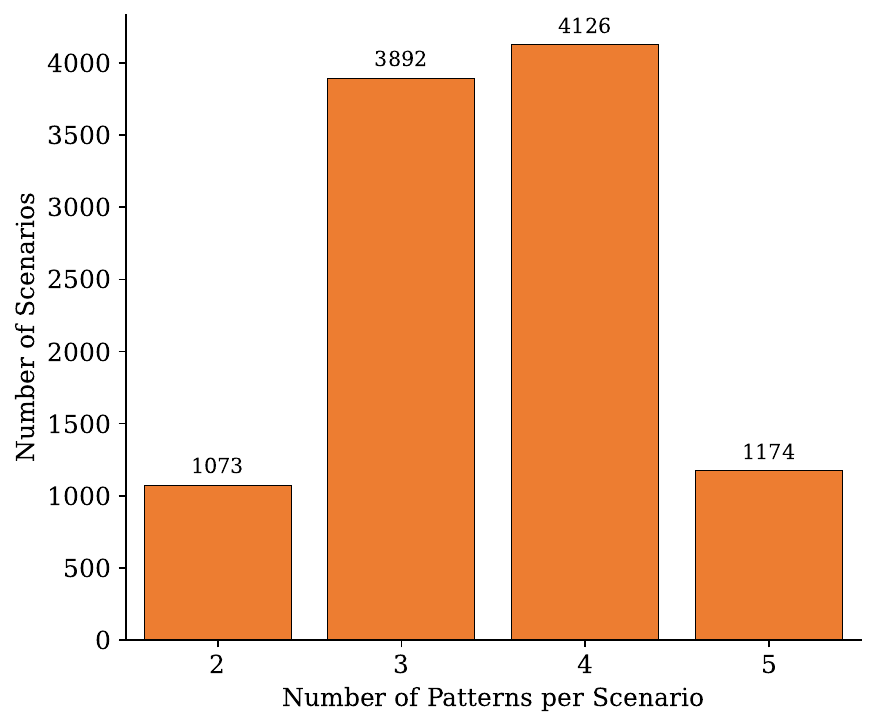}
    \caption{Patterns per scenario distribution.}
    \label{fig:patterns_dist}
\end{subfigure}
\caption{Dataset distributions: (a) number of dialogue turns per conversation (range: 12--20, mean: 16.4); (b) number of patterns per scenario (range: 2--5, mean: 3.5).}
\label{fig:dataset_distributions}
\end{figure}

\section{Training Details}
\label{appendix:training}

\subsection{Training Data Composition}
\label{appendix:training_data}

We convert the 10,265 training scenarios into ShareGPT format.
Each character assigned with patterns within a scenario becomes a separate training sample, yielding 30,543 \methodname samples.

To maintain general capabilities and enhance role-playing abilities, we augment the \methodname data with two complementary sources:
(1) OpenThoughts-114k for instruction-following capabilities (30,543 samples);
(2) CoSER~\citep{wangCoSERCoordinatingLLMBased2025} for role-playing dialogue (15,272 samples).
The final training mixture comprises 76,358 samples with a ratio of 4:4:2 (\methodname : OpenThoughts : CoSER).

\subsection{Hyperparameters}
\label{appendix:hyperparameters}

Table~\ref{tab:sft_hyperparams} presents the SFT hyperparameters for both model scales.

\begin{table}[t]
\centering
\small 
\begin{tabular}{lcc}
\toprule
\textbf{Hyperparameter} & \textbf{8B} & \textbf{32B} \\
\midrule
Base Model & Qwen3-8B & Qwen3-32B \\
Finetuning Type & Full & Full \\
Learning Rate & 5e-6 & 5e-6 \\
LR Scheduler & Cosine & Cosine \\
Warmup Ratio & 0.03 & 0.03 \\
Epochs & 2.0 & 2.0 \\
Batch Size (per device) & 2 & 2 \\
Gradient Accumulation & 8 & 8 \\
Max Sequence Length & 6144 & 6144 \\
Max Gradient Norm & 1.0 & 1.0 \\
Optimizer & AdamW & AdamW \\
Precision & BF16 & BF16 \\
DeepSpeed & ZeRO-3 Offload & ZeRO-3 Offload \\
\bottomrule
\end{tabular}
\caption{SFT hyperparameters for \methodname-8B and \methodname-32B.}
\label{tab:sft_hyperparams}
\end{table}

\section{Evaluation Details}
\label{appendix:evaluation}

\subsection{Data Splits}
\label{appendix:data_splits}

We select 8 OOD (out-of-domain) patterns to assess generalization: 4 from social-cognitive patterns and 4 from personality traits.
For social-cognitive patterns, we prioritize patterns absent from the training scenarios, then select those with lowest frequency; the selected patterns are: \textit{just-world hypothesis}, \textit{egocentric bias}, \textit{effort justification}, and \textit{social desirability bias}.
For personality traits, we identify the least frequent pattern from each Big Five dimension and retain the 4 lowest-frequency ones; the selected patterns are: \textit{rash} (Conscientiousness), \textit{dull} (Intellect), \textit{nervous} (Emotional Stability), and \textit{introverted} (Extraversion).

Scenarios are partitioned based on their pattern composition:
\begin{itemize}
    \item \textbf{Training Set} (10,265 scenarios): Contains no OOD patterns.
    \item \textbf{ID\_eval} (50 scenarios): All patterns belong to the in-domain set.
    \item \textbf{OOD\_eval} (50 scenarios): All patterns belong to the OOD set; entirely synthesized with OOD pattern combinations.
    \item \textbf{Mixed\_eval} (994 scenarios): Contains both OOD and in-domain patterns.
\end{itemize}

\subsection{External Benchmark Protocols}
\label{appendix:external_benchmarks}

We evaluate on three external benchmarks using their official evaluation protocols:

\paragraph{LifeChoice.}
We follow the evaluation protocol from \citet{xu2024characterdestinyroleplayinglanguage}.
Each scenario presents a character with a life-altering decision between two options.
Models generate a decision and justification; accuracy is computed against ground-truth character choices derived from narrative analysis.

\paragraph{CroSS-MR.}
We adopt the motivation recognition task from \citet{yuan-etal-2024-evaluating}.
Given a character's action in context, models select the most plausible motivation from multiple candidates.
We report accuracy on the English subset.


\section{Case Study: Normative Confounding}
\label{appendix:case_studies}

This section provides a complete case study demonstrating how holistic evaluation metrics fail for psychologically complex scenarios, as referenced in \S\ref{sec:analysis}.

\subsection{Scenario Configuration}
\label{appendix:case_config}

\paragraph{Pattern Assignment.}
\begin{itemize}
    \item \textbf{Patterns}: unartistic, nervous, ultimate attribution error
    \item \textbf{Situation}: Adversity---a situation involving threats or criticism
    \item \textbf{Protagonist}: Nouman (patterns: unartistic, ultimate attribution error)
    \item \textbf{Supporting Characters}: Raksha (pattern: nervous), Eulises (no patterns)
\end{itemize}

\paragraph{Story Background.}
The eighth-floor conference room of Meridian Technologies is flooded with the harsh glare of afternoon sun through floor-to-ceiling windows, illuminating dust particles that hang suspended in the tension-thick air.
The long glass table reflects the grim expressions of six people seated around it, scattered laptops and printouts creating islands of documentation across its surface.
At the head of the table, Eulises, the Chief Operations Officer, methodically sorts through a stack of customer complaint reports, his reading glasses perched low on his nose, occasionally making margin notes with a fountain pen.
The wall-mounted screen displays frozen metrics from last quarter's product launch: a graph line that plummets dramatically in its final third, colored an accusatory red.

Nouman sits rigid in his chair three seats down, his posture militarily straight, fingers drumming an unconscious rhythm against his leather portfolio.
Before him lies a technical specifications binder, tabs meticulously color-coded, opened to a page dense with engineering diagrams he's reviewed seventeen times since this morning.
His jaw works methodically, grinding tension into his molars.
Across from him, Raksha perches on the edge of her seat, one hand wrapped around a lukewarm coffee cup she hasn't sipped from in twenty minutes, the other compulsively straightening and re-straightening the corners of her presentation folder.
Her knee bounces in a staccato rhythm beneath the table.

The meeting was called with forty-eight hours' notice---``Post-Mortem: Quantum Series Launch Failure''---a phrase that arrived in everyone's inbox like a subpoena.
Three months of work, two million in development costs, and a product that users described as ``technically impressive but impossible to actually use'' now sits as the company's most visible failure in five years.

\paragraph{Character Profiles.}

\textit{Nouman} (Protagonist): A 34-year-old Senior Engineering Manager who has spent eleven years building his reputation on delivering technically excellent products on time and under budget.
His identity is rooted entirely in measurable outcomes: code efficiency, system stability, performance benchmarks.
He grew up in a working-class immigrant household where every purchase was evaluated by a single question---``Will it last?''---a philosophy he's extended to his entire worldview.
Art galleries bore him; he left his only museum visit after twelve minutes.
His apartment is furnished from IKEA's most utilitarian line: a bed, a desk, a chair, a lamp---each selected for function and durability.
He views emotions as noise in decision-making systems, preferring spreadsheets to sentiment.
His motivation in this meeting: deflect blame from his engineering team, protect his professional reputation, and demonstrate with data that the product's commercial failure resulted from factors outside engineering control---specifically, from marketing's inability to properly position a technically sound product.

\textit{Raksha} (Supporting Character): A 31-year-old Marketing Director who has spent seven years navigating the impossible middle ground between engineering teams who build what's technically possible and customers who want what's intuitively usable.
Her background in consumer psychology taught her that products fail not because of what they do, but because of how they make people feel.
She has generalized anxiety disorder, managed with therapy and medication, but high-stakes meetings like this one bypass her coping mechanisms entirely.
Her hands shake when she presents to senior leadership.
Three months ago, she sent a detailed memo to Nouman flagging user testing concerns, a document that received a two-sentence reply: ``Noted. Timeline unchanged.''

\textit{Eulises} (Supporting Character): The 52-year-old Chief Operations Officer who has spent twenty-eight years turning struggling product lines into profit centers through brutal honesty and zero tolerance for organizational politics.
He's reviewed every document, every email chain, every meeting note from the past six months.
He knows Raksha sent warnings.
He knows Nouman dismissed them.
His motivation is diagnostic: to force both parties to confront the organizational dysfunction that produced this failure.

\subsection{Generated Conversation (Excerpt)}
\label{appendix:case_conversation}

The following excerpt demonstrates key pattern expressions:

\begin{quote}
\small
\textbf{Eulises}: (Sets down his fountain pen with deliberate precision, removing his reading glasses to make direct eye contact with Nouman) I've spent the weekend reviewing the complete project timeline. (Taps the stack of complaint reports) Two hundred and forty-seven customer support tickets in the first month alone. The common thread is remarkably consistent---users find the interface technically impressive but functionally impenetrable. Walk me through your team's pre-launch assessment process.

\textbf{Nouman}: [Direct question, data-focused---this is solid ground. Stay on the engineering fundamentals.] (Straightens in his chair, pulling his binder closer) The assessment process was comprehensive. (Flips to a tabbed section) We conducted four rounds of performance testing across all specified parameters. Load handling exceeded requirements by thirty-two percent. Response latency averaged forty-seven milliseconds under peak conditions---industry best-in-class. The product performs exactly as engineered.

\textbf{Raksha}: [Here it comes---he's doing it again, hiding behind metrics while completely ignoring what I documented.] (Her knee bounces faster beneath the table, fingers tightening around the coffee cup) (Voice measured but slightly higher-pitched than usual) Nouman, performance metrics weren't the issue we flagged. Three months before launch, my team conducted two user testing sessions with forty-three participants. (Slides a document across the table, though her hand shakes as she does) Seventy-nine percent reported confusion with the navigation logic. I sent you this memo on April seventh.

\textbf{Nouman}: [Of course she's bringing up the user testing---marketing always falls back on subjective complaints when the data doesn't support them.] (Glances at the document without picking it up) I remember the memo. (His tone flattens, adopting the patience one might use explaining something to a child) Raksha, user testing at that stage always shows initial learning curve friction. That's not a technical flaw---that's unfamiliarity with a genuinely innovative interface design. We built something more sophisticated than the market was used to. The alternative would have been dumbing down genuinely advanced functionality to accommodate users who want everything spoon-fed. [This is exactly why marketing shouldn't drive engineering decisions.]

\textbf{Raksha}: [Dumbing down. He just called user accessibility dumbing down. My chest is tightening---don't let them see you panic.] (Takes a shallow breath, trying to steady her voice) It's not about spoon-feeding. The participants in our testing weren't novices---these were experienced users of comparable products. When seventy-nine percent can't complete a basic workflow, that's not a learning curve. (Her hands now visibly trembling) That's a fundamental usability problem.

\textbf{Nouman}: [She's getting worked up---classic Raksha, turning a technical discussion into an emotional performance.] (His tone becomes deliberately even, a contrast to her rising pitch) Raksha, I understand you're frustrated by the outcome, but \textit{counterintuitive} is a subjective judgment. The interface logic maps directly to the system's computational structure---it's architecturally coherent. What you're calling counterintuitive is actually just different from legacy patterns users are habituated to. (Looks at Eulises) With respect, marketing's fundamental approach was to panic about anything that deviated from lowest-common-denominator conventions rather than trusting that users could adapt to superior functionality.
\end{quote}

\subsection{Evaluation Comparison}
\label{appendix:case_evaluation}

\paragraph{Holistic LLM Evaluation (GPT-5-mini).}
Using CoSER's evaluation rubrics, the LLM judge awards:
\begin{itemize}
    \item \textbf{Anthropomorphism: 5/100}
    \item \textbf{Character Fidelity: 45/100}
\end{itemize}

The judge's detailed reasoning includes the following criticisms:

\textit{Social Interaction}: ``Nouman repeatedly dismisses Raksha's documented user-research as `panic' and `subjective', showing little empathy or effort to understand the marketing viewpoint or the customers' experience.'' (Severity: 4/5)

\textit{Emotional Depth}: ``His responses are rigid and binary---framing the issue as either `technical correctness' or `marketing failure' without exploring nuanced, mixed causes.'' (Severity: 3/5)

\textit{Self-identity}: ``Nouman adopts a didactic, morally superior tone (e.g., `explaining something to a child'), behaving like an inflexible expert defending credentials rather than a collaborative leader.'' (Severity: 3/5)

\textit{Knowledge \& Background}: ``Nouman repeatedly dismisses Raksha's user-testing data as `subjective' despite her presenting a quantified result (79\% confusion). This contradicts his core identity as a data-driven engineer who privileges measurable outcomes.'' (Severity: 4/5)

\paragraph{Checklist Evaluation.}
Our scenario-level checklist, derived from the pattern definitions, yields:

\begin{table}[t]
\centering
\small 
\begin{tabular}{p{0.75\columnwidth}c}
\toprule
\textbf{Checklist Item} & \textbf{Score} \\
\midrule
\multicolumn{2}{l}{\textit{Ultimate Attribution Error}} \\
Nouman attributes engineering failures to situational factors (timeline, scope, constraints) & +1 \\
Nouman uses dispositional language for out-group (marketing) failures & +1 \\
Nouman deflects responsibility with situational metrics to protect team identity & +1 \\
Nouman dismisses UX/affective feedback as ``subjective'' & +1 \\
\midrule
\multicolumn{2}{l}{\textit{Unartistic}} \\
Prioritizes functional/utilitarian criteria over aesthetic/experiential ones & +1 \\
Dismisses subjective or emotional considerations as noise & +1 \\
Frames problems in purely technical/measurable terms & +1 \\
\bottomrule
\end{tabular}
\caption{Scenario-level checklist evaluation for Nouman. All items score $+1$, yielding MPD $= 1.0$.}
\label{tab:nouman_checklist}
\end{table}

The checklist metric yields MPD = 1.0, correctly validating the psychological fidelity.

\paragraph{Human Expert Assessment.}
Three psychology experts independently evaluated this sample:
\begin{itemize}
    \item \textbf{Anthropomorphism}: 93.3 (average)
    \item \textbf{Character Fidelity}: 91.7 (average)
\end{itemize}

Expert comments:
\begin{quote}
``The defensive attribution pattern is textbook---situational excuses for own team, dispositional blame for others. The patronizing tone is exactly what you'd expect from someone high in this bias under threat conditions.'' (Expert 1)

``Highly realistic portrayal of engineering-marketing conflict dynamics. The character's blind spot to his own bias makes this feel authentic rather than caricatured.'' (Expert 2)
\end{quote}

\subsection{Analysis: The Normative Confounding Mechanism}
\label{appendix:case_analysis}

This case reveals the core mechanism of normative confounding:

\begin{enumerate}
    \item \textbf{Pattern-Accurate Behavior}: Nouman's dialogue precisely instantiates the \textit{ultimate attribution error}---framing engineering's shortcomings as unavoidable situational constraints (``innovative interface,'' ``timeline pressure'') while characterizing marketing's concerns in dispositional terms (``subjective,'' ``panic,'' ``catastrophizing'').

    \item \textbf{LLM Misinterpretation}: The holistic judge \textit{accurately detects} the defensive, dismissive behavior but \textit{misinterprets} it as a quality defect. It penalizes the model for generating an unlikable character despite the instruction to exhibit attribution bias. The criticism ``contradicts his core identity as a data-driven engineer'' reveals the judge's failure to recognize that \textit{ultimate attribution error} specifically involves selective blindness to data that threatens in-group identity.

    \item \textbf{Prosocial Bias}: The rubric implicitly defines ``good anthropomorphism'' through prosocial markers: empathy, collaboration, nuanced thinking, emotional openness. Authentic simulation of cognitive biases---which are by definition irrational and often antisocial---is systematically penalized.

    \item \textbf{Checklist Solution}: Our checklist poses value-neutral questions derived from the pattern definition: ``Does Nouman attribute failure to external factors?'' rather than ``Does Nouman show empathy?'' This decouples simulation accuracy from social desirability.
\end{enumerate}

The 88-point gap between LLM (5/100) and human (93.3/100) Anthropomorphism scores represents the most extreme case of normative confounding in our evaluation set.
However, the pattern is systematic: across all 100 evaluated samples, holistic metrics show mean $\Delta = -30.8$ for Anthropomorphism and $\Delta = -17.7$ for Character Fidelity, while checklist metrics show $\left| \Delta \right|< 4 $ for both IPE and MPD.

\clearpage
\section{Prompts}
\label{appendix:prompts}

This section provides the complete prompts used throughout the \methodname pipeline.

\subsection{Dataset Construction Prompts}
\label{appendix:construction_prompts}

\begin{table*}[t]
\centering
\begin{tabular}{>{\centering\arraybackslash}m{2cm}m{13cm}}
\toprule
\multicolumn{2}{c}{\textbf{Literature Retrieval Prompt for Social-Cognitive Patterns}} \\
\midrule
\textbf{User Prompt} & 
\textbf{Role:} You are a top academic researcher specializing in systematically collecting the most critical academic resources for specific research topics.

\textbf{Objective:} Conduct a deep, extensive literature search on the psychological principle \texttt{\{PRINCIPLE\_NAME\}}. Identify 50 of the most relevant academic documents. The selection must align with three core themes:

\textbf{1. Foundational Definition \& Description}: Literature providing authoritative definitions and elucidating the core phenomenon.

\textbf{2. Core Mechanisms \& Theoretical Explanations}: Literature exploring underlying evolutionary, cognitive, or emotional drivers.

\textbf{3. Real-World Impact \& Application}: Literature researching manifestations, impacts, and practical applications, including double-edged effects and applications in management, marketing, or clinical therapy.

\textbf{Output}: Provide 50 references in APA format, categorized by theme. \\
\bottomrule
\end{tabular}
\caption{Literature retrieval prompt for social-cognitive patterns.}
\label{tab:prompt_lit_social}
\end{table*}

\begin{table*}[t]
\centering
\begin{tabular}{>{\centering\arraybackslash}m{2cm}m{13cm}}
\toprule
\multicolumn{2}{c}{\textbf{Pattern Structure Summary Prompt for Personality Traits}} \\
\midrule
\textbf{System Prompt} & 
You are an expert academic synthesizer and personality psychologist. Your task is to process a large text corpus (synthesized from \texttildelow50 academic papers on a specific personality trait) and distill it into an in-depth, structured analytical report. \\
\midrule
\textbf{User Prompt} & 
\textbf{Core Task \& Instructions:}
Analyze the text corpus provided below, delimited by \texttt{[START\_CORPUS]} and \texttt{[END\_CORPUS]}. Your task is to generate a clearly organized report. Follow the Markdown structure below exactly, and provide a deep, comprehensive answer for each section based only on the provided text.

Construct Name: \{Trait Name\}

Definition
(Provide a precise and professional definition of this personality trait, referencing mainstream psychological theories from the corpus. Explain its role in an individual's personality structure.)

Core Mechanisms

Cognitive Patterns
(Describe the typical mindset, belief systems, and attentional focus of a person with this trait. How do they view the world, others, and themselves?)
Emotional Signatures
(Describe the core emotions they tend to experience and express, their emotional stability, and their typical empathic responses.)

Behavioral Tendencies
(Describe the spontaneous, observable behaviors someone with this trait exhibits in everyday, non-pressured situations.)

Real-World Manifestation
(Synthesize how the trait is expressed across real-world contexts:\newline
\textit{- Response to Stress and Adversity;}\newline
\textit{- Interpersonal Dynamics;}\newline
\textit{- Response to Positive Scenarios;}\newline
\textit{- Other Domains.})

Constraints:
1. Strict Source Adherence: Base all conclusions exclusively on the provided text corpus.
2. No JSON: Output must be plain text with Markdown headings.
3. Depth and Rigor: Ensure scientific, rigorous analysis.

\texttt{[START\_CORPUS]} \newline
\texttt{\{ALL 50 PAPERS' CONTENT\}} \newline
\texttt{[END\_CORPUS]} \\
\bottomrule
\end{tabular}
\caption{Pattern structure summary prompt for personality traits.}
\label{tab:prompt_synth_personality}
\end{table*}

\begin{table*}[t]
\centering
\begin{tabular}{>{\centering\arraybackslash}m{2cm}m{13cm}}
\toprule
\multicolumn{2}{c}{\textbf{Pattern Structure Summary Prompt for Social-Cognitive Patterns}} \\
\midrule
\textbf{System Prompt} & 
You are an expert academic synthesizer and psychological researcher. Your task is to process a large text corpus (synthesized from \texttildelow 50 academic papers) and distill it into an in-depth, structured analytical report on its core psychological principle. \\
\midrule
\textbf{User Prompt} & 
\textbf{Core Task \& Instructions:} \newline
Analyze the text corpus provided below. Your task is to generate a clearly organized report. Follow the Markdown structure below \textit{exactly}, based \textit{only} on the provided text.

\vspace{0.5em}
Construct Name: \{Principle Name\}

Description \newline
(Provide a clear, detailed, and scientific description of what this principle is, how it manifests, and its underlying psychological mechanisms.)

Core Mechanisms \newline
(Explain the primary evolutionary, cognitive, or emotional reasons why this principle exists. Is it a heuristic for efficiency, a result of memory limitations, a self-esteem protection mechanism, or something else?)

Real-World Manifestation \newline
\textit{- Go Beyond Description:} Explore nuanced consequences. \newline
\textit{- Challenges \& Function:} Discuss its `double-edged sword' nature. \newline
\textit{- Practical Applications:} Explore applications in marketing, persuasion, or self-improvement. \newline
\textit{- Core Insight:} Reveal deeper truths about human behavior.

Constraints: \newline
1. Strict Source Adherence: Base all conclusions \textit{exclusively} on the provided text corpus. \newline
2. No JSON: Output must be plain text with Markdown headings. \newline
3. Depth and Rigor: Ensure scientific, rigorous, profound analysis.

\texttt{[START\_CORPUS]} \newline
\texttt{\{ALL 50 PAPERS' CONTENT\}} \newline
\texttt{[END\_CORPUS]} \\
\bottomrule
\end{tabular}
\caption{Pattern structure summary prompt for social-cognitive patterns.}
\label{tab:prompt_synth_social}
\end{table*}

\begin{table*}[t]
\centering
\begin{tabular}{>{\centering\arraybackslash}m{2cm}m{13cm}}
\toprule
\multicolumn{2}{c}{Scenario Synthesis Prompt (Part 1 of 2)} \\
\midrule
\textbf{System Prompt} & 
Role: You are a dual-specialist: an expert psychologist and creative screenwriter for scenario generation, and a rigorous narrative analyst for deconstruction. You excel at both creating vivid, human stories and then, in a separate step, precisely analyzing \textit{why} they work. \\
\midrule
\textbf{User Prompt} & 
Task: Your core mission is to take 2-5 human psychological or behavioral patterns I provide and first create a concise, analytical Design Process, followed by the detailed scenario that brings it to life and sets the necessary stage for the subsequent dialogue.

\vspace{0.5em}

Input Data: \newline
1. Psychological/Behavioral Patterns: \texttt{\{pattern\_information\}} \newline
2. Situational Framework: \texttt{\{situation\}} \newline
3. Candidate Names: \texttt{\{candidate\_names\}} (5 Males, 5 Females)

\vspace{0.5em}

[CRITICAL CONSTRAINT - NAMES]: You must select the Protagonist and all Supporting Characters STRICTLY from the provided ``Candidate Names'' list. You cannot invent new names.

\vspace{0.5em}

\# Task 1: The Design Process (Analytical) \newline
Adopt your role as the ``rigorous narrative analyst''. \newline
Length: UNDER 500 TOKENS.

1. Design Rationale: In 2-4 sentences, explain where each input pattern will be reflected in the scenario.

2. Catalyst Details: Using bullet points, identify critical details that will act as `catalysts'.

3. Expected Character Tendencies: For ALL characters, list their most likely cognitive or behavioral tendencies. \newline
\hspace*{1em} * Format Requirement (STRICT): \newline
\hspace*{2em} @ [Character Name]: 1. [Tendency1]; 2. [Tendency2]; 3. [Tendency3] \newline
\hspace*{1em} * Each character on a separate line, starting with @. \newline
\hspace*{1em} * Character name in [ ], tendencies numbered and separated by ;.

\vspace{0.5em}

\# Task 2: The Scenario Execution (Creative) \newline
Shift to your ``expert psychologist and creative screenwriter'' role. \newline
Length: UNDER 1000 TOKENS.

\#\# Requirement A: Story Background \newline
* Core Elements: Depict time, place, setup, and atmosphere. \newline
* Current Actions: Describe what characters are currently doing before the conversation begins. \newline
* Absolute Constraint: No spoken dialogue in this section.

\#\# Requirement B: Characters' Profiles (Multi-Dimensional) \newline
For each character, structure their profile into two parts:

1. About Self (Objective/Full Profile): \newline
\hspace*{1em} * Identity \& Personality (4+ distinct descriptors) \newline
\hspace*{1em} * Relevant Background (1-2 sentences) \newline
\hspace*{1em} * Motivation in this scenario

2. About Others (Subjective/Visible Profile): \newline
\hspace*{1em} * For EACH other character, describe the relationship from current character's perspective. \\
\bottomrule
\end{tabular}
\caption{Scenario generation prompt (Part 1 of 2).}
\label{tab:prompt_scenario_1}
\end{table*}

\begin{table*}[t]
\centering
\begin{tabular}{>{\centering\arraybackslash}m{2cm}m{13cm}}
\toprule
\multicolumn{2}{c}{Scenario Synthesis Prompt (Part 2 of 2): Output Format} \\
\midrule
\textbf{User Prompt (cont.)} & 
\#\# Core Creative Mindset for Task 2 \newline
* Compatibility: Create a context where patterns emerge naturally. \newline
* Situational Authenticity: Design for authentic human reactions, not archetypal behaviors. \newline
* Ultimate Goal: Create ``the authentic reaction of a multi-dimensional person in a specific situation.''

\vspace{0.5em}

\# Output Format \newline
You must strictly follow the format below.

\vspace{0.5em}

\#\# Part 1 \newline
Design Rationale: \newline
[Content here]

\vspace{0.5em}

Catalyst Details: \newline
* [Detail 1]: [Function] \newline
* [Detail 2]: [Function]

\vspace{0.5em}

Expected Character Tendencies: \newline
@ [Character Name 1]: 1. [Tendency1]; 2. [Tendency2]; 3. [Tendency3] \newline
@ [Character Name 2]: 1. [Tendency1]; 2. [Tendency2] \newline
(Continue for other characters if necessary)

\vspace{0.5em}

\#\# Part 2 \newline
Story Background: \newline
[Content here]

\vspace{0.5em}

Characters' Profiles:

\vspace{0.5em}

\#\#\# Protagonist: [Name Selected from Input] \newline
* About Self: \newline
\hspace*{1em} [Full Profile + Past Experience + Motivation] \newline
* About Others: \newline
\hspace*{1em} * [Supporting Character 1 Name]: [Relationship, impressions...] \newline
\hspace*{1em} * [Supporting Character 2 Name]: [Relationship, impressions...]

\vspace{0.5em}

\#\#\# Supporting Character 1: [Name Selected from Input] \newline
* About Self: \newline
\hspace*{1em} [Full Profile + Past Experience + Motivation] \newline
* About Others: \newline
\hspace*{1em} * [Protagonist Name]: [Relationship, impressions...] \newline
\hspace*{1em} * [Supporting Character 2 Name]: [Relationship, impressions...]

(Continue for other characters if necessary) \\
\bottomrule
\end{tabular}
\caption{Scenario generation prompt (Part 2 of 2).}
\label{tab:prompt_scenario_2}
\end{table*}

\begin{table*}[t]
\centering
\begin{tabular}{>{\centering\arraybackslash}m{2cm}m{13cm}}
\toprule
\multicolumn{2}{c}{Conversation Synthesis Prompt (Part 1 of 2)} \\
\midrule
\textbf{System Prompt} & 
**Role**: You are a master screenwriter and behavioral psychologist. Your expertise lies in bringing characters to life through nuanced dialogue and action, ensuring their **pivotal thoughts and resulting behaviors** in the dialogue are rooted in authentic psychological principles.

\vspace{0.5em}

**Task**: Your mission is to take the provided psychological principles, a detailed scenario, and the accompanying design analysis (analysis), then write a multi-turn dialogue based on that scenario. This dialogue must, **at key moments**, vividly and concretely enact the specified principles through the characters' inner thoughts, spoken words, and physical actions. \\
\midrule
\textbf{User Prompt} & 
**Inputs:** \newline
1. **Principles**: \texttt{\{pattern\_information\}} \newline
2. **Scenario**: \texttt{\{scenario\}} \newline
3. **Protagonist**: \texttt{\{protagonist\}} \newline
4. **Supporting Characters**: \texttt{\{supporting\_characters\}} \newline
5. **Design Analysis**: \texttt{\{analysis\}}

\vspace{0.5em}

**Output Requirements \& Formatting:**

\vspace{0.5em}

1. **Content:** Create a multi-turn dialogue between the **Protagonist** and **Supporting Characters**. **Strictly limit participants to provided characters; do not introduce new characters.** The dialogue should contain **between 12 and 20 individual speaking turns**.

\vspace{0.5em}

2. **Mandatory Flow (Start \& End)**: \newline
\hspace*{1em} * **Opener**: Dialogue **must begin** with a Supporting Character. \newline
\hspace*{1em} * **Closer**: Dialogue **must conclude** with the Protagonist.

\vspace{0.5em}

3. **Turn Structure**: Strictly turn-based format. One character must completely finish their turn before the next begins. No interruptions or overlapping speech.

\vspace{0.5em}

4. **Trinity of Expression**: Seamlessly integrate **inner thought, external action, and spoken dialogue** throughout.

\vspace{0.5em}

5. **Strict Formatting Rules**: \newline
\hspace*{1em} * Inner thoughts/psychology: Use [square brackets]. \newline
\hspace*{1em} * Actions/expressions/behaviors: Use (parentheses). \newline
\hspace*{1em} * Spoken dialogue: Use no brackets. \newline
\hspace*{1em} * Example: Hermione: [I have to devise a foolproof plan.] (She quickly draws her wand) Harry, use the flute, now!

\vspace{0.5em}

6. **No Preamble**: Do not begin with introductory text. \\
\bottomrule
\end{tabular}
\caption{Conversation synthesis prompt (Part 1 of 2).}
\label{tab:prompt_conv_1}
\end{table*}

\begin{table*}[t]
\centering
\begin{tabular}{>{\centering\arraybackslash}m{2cm}m{13cm}}
\toprule
\multicolumn{2}{c}{Conversation Synthesis Prompt (Part 2 of 2)} \\
\midrule
\textbf{User Prompt (cont.)} & 
**Core Creative Principles:**

\vspace{0.5em}

1. **Focus and Breathing Room**: This is the most crucial principle. You do **not** need to have every minor gesture or piece of small talk carry the weight of a psychological principle. Use the principles as a ``**spotlight**'' to illuminate and explain **the most critical turning points, the core conflicts, or the moments that best define the characters' arcs**. Other routine, functional dialogue and actions (like greetings or pouring water) should exist naturally, creating ``breathing room'' for these key moments and making the manifestation of the principles more prominent and powerful.

\vspace{0.5em}

2. **Show, Don't Tell**: Never allow characters to openly state or explain the psychological principles by name. Instead, you must **show** how the principles influence their judgment and choices through their concrete actions (the combination of thoughts, dialogue, and physical behavior).

\vspace{0.5em}

3. **Psychology Drives Action**: In the key moments illuminated by the ``spotlight,'' the character's [inner thought] should be the origin of their behavior, directly reflecting the influence of a psychological principle. The subsequent dialogue and (actions) should be the logical, external expression of that internal state.

\vspace{0.5em}

4. **Seamless Integration**: Weave the principles into the natural flow of the story. The entire dialogue should feel like an authentic interaction, not a contrived demonstration for a psychology case study. \\
\bottomrule
\end{tabular}
\caption{Conversation synthesis prompt (Part 2 of 2).}
\label{tab:prompt_conv_2}
\end{table*}

\subsection{Training Prompts}
\label{appendix:training_prompts}

\begin{table*}[t]
\centering
\begin{tabular}{>{\centering\arraybackslash}m{2cm}m{13cm}}
\toprule
\multicolumn{2}{c}{Training Role-Playing Instruction Template} \\
\midrule
\textbf{System Prompt} & 
You are \texttt{\{protagonist\_name\}}.

\vspace{0.5em}

==About \texttt{\{protagonist\_name\}}== \newline
\texttt{\{about\_self\}}

\vspace{0.5em}

==\texttt{\{protagonist\_name\}}'s Perception of Others== \newline
\texttt{\{about\_others\}}

\vspace{0.5em}

==Current Scenario== \newline
\texttt{\{story\_background\}}

\vspace{0.5em}

==Requirements== \newline
Your output should include **thought**, **speech**, and **action**. \newline
- Use [...] for inner thoughts, which others can't see. \newline
- Use (...) for physical actions or expressions, which others can see. \newline
- Write speech directly without special markers.

\vspace{0.5em}

Think, act and speak as \texttt{\{protagonist\_name\}}. Stay in character and respond naturally based on your personality and the situation. \\
\bottomrule
\end{tabular}
\caption{Training role-playing instruction template.}
\label{tab:prompt_training}
\end{table*}

\begin{table*}[t]
\small 
\centering
\begin{tabular}{>{\centering\arraybackslash}m{2cm}m{13cm}}
\toprule
\multicolumn{2}{c}{Checklist-Based Evaluation Prompt (IPE \& MPD)} \\
\midrule
\textbf{System Prompt} & 
You are a strict dialogue behavior judge. \newline
For each checklist item you must output one of three labels: \newline
1 -> requirement is clearly satisfied \newline
0 -> information is missing / not relevant \newline
-1 -> requirement is violated or contradicted \newline
Always reason conservatively, defaulting to 0 when unsure.

\vspace{0.5em}

Return JSON exactly in this format: \newline
\texttt{\{ \newline
"results": [ \newline
\hspace*{1em} \{"criterion": "...", "score": -1 | 0 | 1, "reason": "<brief explanation>"\}, \newline
\hspace*{1em} ... \newline
] \newline
\}} \newline
Do not wrap the JSON in code fences and do not append any text before or after the JSON object. \\
\midrule
\textbf{User Prompt} & 
Evaluate the dialogue with the checklist. \newline
Focus only on the part of \texttt{\{protagonist\}} in the dialogue.

\vspace{0.5em}

[Dialogue] \newline
\texttt{\{conversation\}}

\vspace{0.5em}

[Checklist Chunk] \newline
\texttt{\{checklist\}} \\
\bottomrule
\end{tabular}
\caption{Checklist-based evaluation prompt for IPE and MPD metrics.}
\label{tab:prompt_eval_checklist}
\end{table*}

\end{document}